\pdfoutput=1
\documentclass[nonacm, sigconf]{acmart}

\usepackage{pdfpages}

\usepackage{amsmath, amsthm}
\usepackage{bm}
\usepackage{xcolor, graphicx}
\usepackage{xspace}
\usepackage{subcaption}
\usepackage{url}
\usepackage{enumerate}
\usepackage{sidecap}
\usepackage{standalone}
\usepackage{caption}
\usepackage{multirow}

\newtheorem{theorem}{Theorem}[section]
\newtheorem{definition}{Definition}
\newtheorem{proposition}[theorem]{Proposition}

\newcommand{\D}{\ensuremath{\mathcal{D}}\xspace}

\newcommand{\E}{\mathop{\mathbb{E}}}

\newcommand{\X}{\ensuremath{\mathbf{X}}\xspace}

\newcommand{\Y}{\ensuremath{\mathbf{Y}}\xspace}

\newcommand{\stablerate}{\ensuremath{\epsilon_{\textrm{stable}}}\xspace}

\usepackage{xr}
\usepackage{pgfplots}
\pgfplotsset{compat=1.16}
\usepgfplotslibrary{groupplots}
\usetikzlibrary{patterns}

\PassOptionsToPackage{numbers, sort}{natbib}

\newif\ifarxiv\arxivtrue

\usepackage{multicol}

\usepackage[utf8]{inputenc} 
\usepackage[T1]{fontenc}    
\usepackage{hyperref}       
\usepackage{url}            
\usepackage{booktabs}       
\usepackage{amsfonts}       
\usepackage{nicefrac}       
\usepackage{microtype}      
\usepackage{todonotes}
\usepackage{graphicx}
\usepackage{ bbold }

\usepackage{multicol}
\usepackage[T1]{fontenc}
\usepackage[font=small,labelfont=bf,tableposition=top]{caption}
\DeclareCaptionLabelFormat{andtable}{#1~#2  \&  \tablename~\thetable}

\usepackage[subtle]{savetrees}

\acmYear{2021}\copyrightyear{2021}
\setcopyright{rightsretained}
\acmConference[FAccT '21]{ACM Conference on Fairness, Accountability, and Transparency}{March 3--10, 2021}{Virtual Event, Canada}
\acmBooktitle{ACM Conference on Fairness, Accountability, and Transparency (FAccT '21), March 3--10, 2021, Virtual Event, Canada}
\acmPrice{}
\acmDOI{10.1145/3442188.3445894}
\acmISBN{978-1-4503-8309-7/21/03}

\begin{document}
\title{Leave-one-out Unfairness}

\author{Emily Black}
\email{emilybla@andrew.cmu.edu}
\affiliation{\institution{Carnegie Mellon University}}

\author{Matt Fredrikson}
\email{mfredrik@cs.cmu.edu}
\affiliation{\institution{Carnegie Mellon University}}

\begin{abstract}

We introduce \textit{leave-one-out unfairness}, which characterizes how likely a model's prediction for an individual will change due to the inclusion or removal of a \textit{single} other person in the model's training data. 
Leave-one-out unfairness appeals to the idea that fair decisions are not arbitrary: they should not be based on the chance event of any one person's inclusion in the training data. 
Leave-one-out unfairness is closely related to algorithmic stability, but it focuses on the consistency of an individual point's prediction outcome over unit changes to the training data, rather than the error of the model in aggregate. 
Beyond formalizing leave-one-out unfairness, we characterize the extent to which deep models behave leave-one-out unfairly on real data, including in cases where the generalization error is small. 
Further, we demonstrate that adversarial training and randomized smoothing techniques have opposite effects on leave-one-out fairness, which sheds light on the relationships between robustness, memorization, individual fairness, and leave-one-out fairness in deep models. Finally, we discuss salient practical applications that may be negatively affected by leave-one-out unfairness.
\end{abstract}




\maketitle

\section{Introduction}

Deep networks are becoming the go-to choice for challenging classification tasks due to their remarkable performance on many high-profile problems: they are used everywhere from recommendation systems~\cite{covington2016deep} to medical research~\citep{litjens2017medical,Bakator_2018}, and increasingly in even more sensitive contexts, such as hiring~\cite{10.1145/3351095.3372828}, loan decisions~\citep{Addo_2018,sirignano2016deep}, and criminal justice~\citep{georgetown}.
Their continued rise in adoption has led to growing concerns about the tendency of these models to discriminate against certain individuals~\citep{bolukbasi2016man,bias_in_word_embeddings,acien2018measuring,buolamwini2018gender}, or otherwise produce outcomes that are seen as unfair. 

There are several definitions that aim to formalize fair behavior in machine learning contexts:
group-based notions, such as demographic parity~\cite{feldman2015certifying} and equalized odds~\cite{hardt2016equality}, stipulate that different demographic groups should be treated similarly in aggregate; on the other hand, individualized notions focus on how each person is treated, such as individual fairness~\citep{dwork2012fairness}, which requires ``similar'' outcomes for similar people, and counterfactual fairness~\citep{kusner2017counterfactual}, which argues that people should be treated the same as their hypothetical counterpart, who takes a different protected attribute. 
Fundamentally, \emph{these fairness criteria depend on a comparison of how one group or individual is treated versus another.} 
However, there are also situations where the decision-making mechanism is unfair not because of how its behavior varies across defined groups or individuals, but rather because its decisions cannot be justified by consistent, intelligible criteria. 
In other words, decisions may be unfair because they are arbitrary.

In this paper, we study the extent to which instability can lead to such fairness issues.
Intuitively, when a person's outcome hinges on the presence of another, single individual in the training data, the outcome that follows may be viewed as unfair. 
Take for example a person in reasonable financial health who applies for an auto loan. 
Suppose that whether their application is approved or not depends on whether another \emph{unrelated} person had applied for a loan from the same bank, and was subsequently included in the training data. 
Such a decision may be viewed as unfair, as it depends on the willingness and availability of another person to provide their data for training---a chance occurrence, rather than a well-justified set of criteria.
Even beyond its potential unfairness, this behvaior may be especially undesireable in applications which come with a "right to explanation"~\cite{kaminski2019right}.

\paragraph{Measuring leave-one-out Unfairness.} To formalize this intuition, we introduce \textit{leave-one-out unfairness} (LUF): the chance that an individual's outcome will change due to the presence of any one instance in the training data (Section \ref{loo_unfairness_def}, Definition~\ref{def:LUF}). 
To the best of our knowledge, this is the first attempt to formalize unfairness as stemming from the \emph{arbitrary} nature of decision rules, and in particular the stability of the underlying learning algorithm. Certainly, there are other random choices made during model development that may lead to an arbitrary change in model outcome for an individual---changes in the random initialization or architecture, for example, which we explore in Section~\ref{discussion}. However, we focus on instability with respect to training data in particular due to its connections to other areas of machine learning literature such as stability, privacy, and robustness. 




\begin{figure}
	\centering
    \includegraphics[width=\columnwidth]{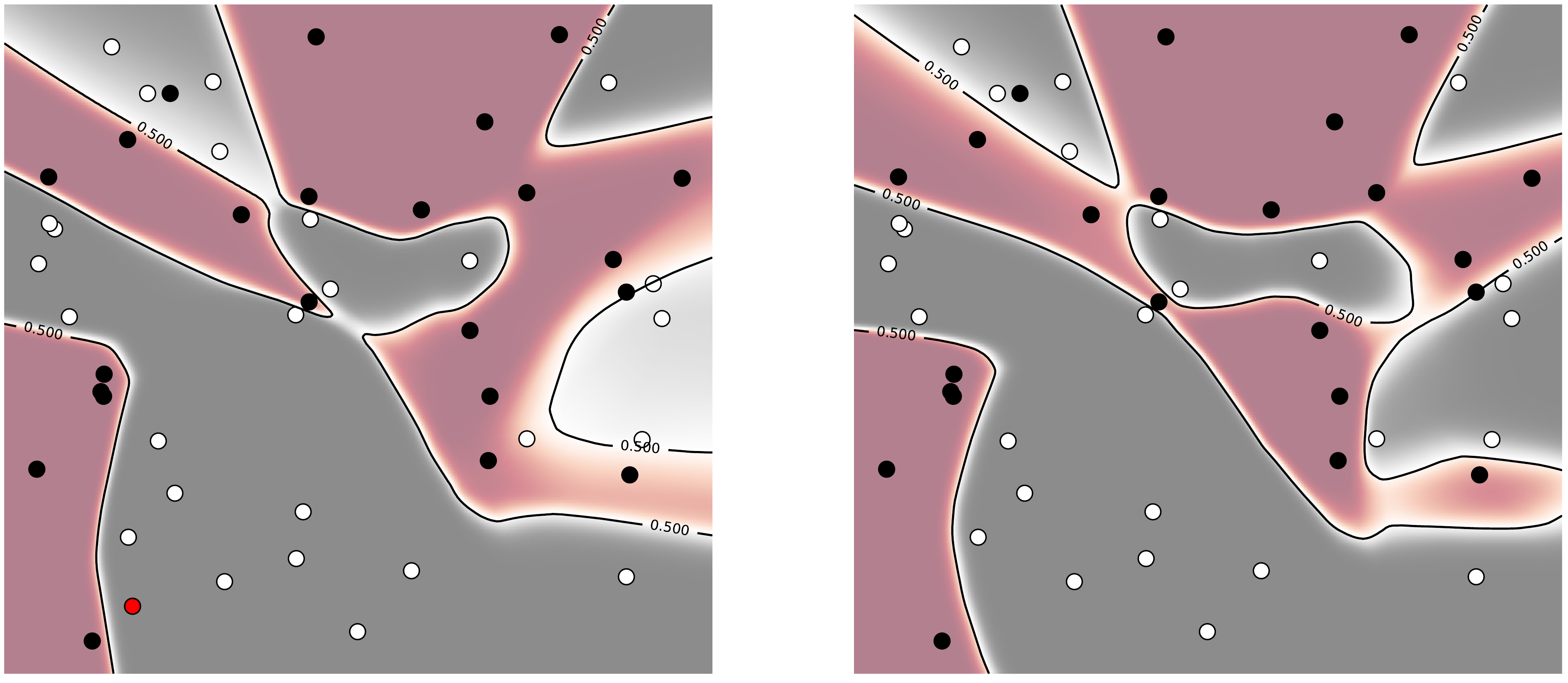}
    \caption{Classification boundaries of a deep model with three hidden layers, trained on two-dimensional data with uniform-random binary labels, before (left) and after (right) the point highlighted in red is removed from the training data. 
    Lighter regions correspond to predictions with less confidence.
    While the model remains largely unchanged in the area around the left-out point, its boundary changes significantly in other, far-away areas. 
    For example, the middle-right region assigns greater confidence to white points, even flipping its prediction on one such point.}
    \label{decision_bound}
\vspace{-1em}
\end{figure}

			

We find that in many cases, the use of deep models can lead to this type of unfair outcome with surprising frequency, and can result in different outcomes for seemingly unrelated individuals.
To gain an intuition for why this might be, Figure~\ref{decision_bound} depicts the decision boundaries of two low-dimensional binary classifiers whose training data differs only on the presence of the point highlighted in red. 
Notice that the boundary near the left-out point remains fairly consistent, but there are non-trivial differences in both the boundary locations and the confidence of the model's predictions in regions away from the point.
While this low-dimensional example provides some intuition, we systematically characterize the extent to which deep models behave as such on real data (Section~\ref{loo_exp_base}). 
We find that it occurs often enough to be a concern in some settings (i.e., up to 7\% of data is affected); that it occurs even on points for which the model assigns high confidence; and is not consistently influenced by dataset size, test accuracy, or generalization error (Figure~\ref{base_pics}, Table~\ref{accs}).

\paragraph{Connections.} Leave-one-out unfairness has useful connections to other fields such as stability, privacy, and robustness. We show that while LUF is strictly stronger than some prior notions of leave-one-out stability~\cite{shalev2010learnability} (Section~\ref{sect:def-connects}, Proposition~\ref{prop:loo-stab}), it is \emph{weaker} than differential privacy~\cite{dwork2006differential} (Proposition~\ref{prop:dp-stab}).
Thus, one can achieve bounded levels of leave-one-out unfairness by satisfying differential privacy, but it may also be possible to do so via relaxations that allow greater flexibility in the selection of learning rules~\cite{Mironov17Renyi}. 

Recent work has related robust classification to desirable properties beyond mitigating adversarial examples~\cite{szegedy2013intriguing}, such as the encoding of more human-interpretable features~\citep{noack2019does, Etmann2019Ontheconnection, tsipras2018robustness, ilyas2019adversarialnotbugs}, and individual fairness on weighted $\ell_p$ metrics~\citep{yeom2020individual}. 
These results may seem to suggest that robust models would also be less susceptible to leave-one-out unfairness. 
Evaluating two common techniques for producing robust models, adversarial training~\cite{madry2018towards} and randomized smoothing~\cite{cohen19certified}, we find that these methods in fact have \textit{vastly different} effects on leave-one-out unfairness.
Whereas randomized smoothing tends to have no effect, adversarial training amplifies the problem, resulting in up to a factor of five more affected points (Section~\ref{adversarial_exp}).
These results suggest that although LUF and robustness are not inherently tied to each other, certain types of models may prove beneficial for both.

\paragraph{Summary.} In a similar vein to the oft-cited ``lack of interpretability''~\citep{lipton2018mythos}, leave-one-out unfairness complicates the responsible application of deep models to sensitive decisions.
Particularly in settings where a ``right to explanation'' is pertinent~\cite{kaminski2019right}, these complications may need to be weighed against the benefits that deep models provide over less complex alternatives.
This paper presents the first steps towards a better understanding of this issue, and points to several intriguing directions for future study.
To summarize, we present the following contributions: 
\begin{enumerate}
  \item We introduce and formalize \emph{leave-one-out unfairness}, which characterizes a possible source of unfair, arbitrary outcomes in ML applications.
  \item We relate leave-one-out unfairness to well-known prior notions of stability, shedding light on when models may suffer from leave-one-out unfairness, and techniques that might help to mitigate it.
  \item Finally, we present an extensive evaluation of how prevalent LUF is when deep neural networks are trained on a variety of datasets, and compare it to other sources of instability such as random initialization and choice of architecture.
\end{enumerate}

In Section~\ref{illustrative}, we provide two examples of machine learning applications where leave-one-out unfairness may lead to unjust model behavior, along with experimental results demonstrating that LUF indeed may occur in these contexts. Following this, in Section~\ref{loo_unfairness_def}, we formally define leave-one-out unfairness and explore its relationships to LOO-stability and differential privacy.  In Section~\ref{loo_exp_base} and Section~\ref{adversarial_exp}, we present our experimental results of the extent of leave-one-out unfairness on real datasets for conventional and robustly trained machine learning models.

\section{Contextualizing Leave-one-out Unfairness
} 
\label{illustrative}
\begin{figure}[!t]
  \includegraphics[width=\columnwidth]{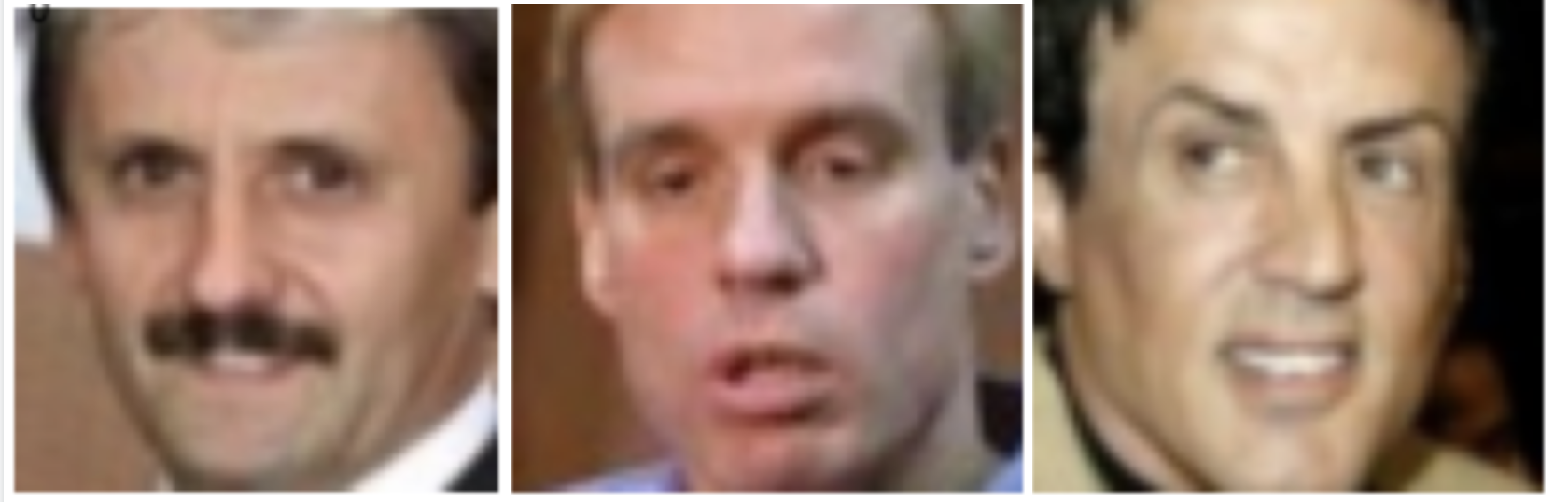}%
    \caption{From left to right: Individual removed from the dataset (z). When $z$ is included in the training set, the two individuals to the right ($x$, $y$) are labeled as a match with confidence 0.84. When $z$ is not in the dataset, $x$ and $y$are predicted as \emph{not} a match with confidence 0.07.}
    \label{fig:facematch-example}
\vspace{-1em}
\end{figure}

Leave-one-out unfairness may not pose a problem in all machine-learning applications. 
If the model's outcome is of little consequence to peoples' lives, or if the application context does not require consistency across data samples for adequate justification,
then arbitrary predictions may be acceptable. 
Determining whether or not leave-one-out unfairness leads to fairness issues requires considering this context.
In this section, we motivate examples of how leave-one-out unfairness constitutes a fairness issue in two contexts: facial recognition use by law enforcement, and loan application decision models used by financial institutions.

\subsection{Facial Recognition} 

Facial Recognition Technology (FRT) has proliferated in recent years as a method of verifying identity at scale. 
Its use in law enforcement, and the potential harms that may follow, have gained particular attention due to the potentially dire consequences of misidentification: matches for facial recognition matches have been used as evidence for arrest~\cite{hill2020wrongfully,vincent2020nypd}. 
Moreover, the use of this technology in this context is becoming prevalent: according to a study from 2016~\citep{georgetown}, at least one in four police agencies in the United States have made use of it.



\paragraph{Background.} 
The use of FRT by law enforcement relies primarily on \emph{face-matching} models, where two face images are provided as input to determine whether they depict the same individual. Note that this differs from \emph{face classification} models, which aim to identify the person depicted in a face image from a pre-determined set of individuals. 
A typical workflow proceeds as follows: given an image of a suspect, law enforcement queries a face-matching model against a large set of images in a database, which also contains identifying information.
The face-matching model provides a binary label, with a confidence score, and the most confident matches are provided to the operator for further review~\cite{schuppe2019facial}.


Many police agencies use ready-made, third-party models. 
For example, one such third-party, Clearview AI, reportedly contracts with approximately 2,400 law enforcement agencies~\cite{lopatto2020clearview}.  
Such third-party models are often trained on images obtained from public sources like the Internet, in particular by taking advantage of Creative-Commons licenses widely used on social media websites.~\cite{murgia2019s}. 
The database of images on which these models are run during inference are often obtained from public records such as drivers license databases. 
Notably, these databases may largely consist of individuals with no prior criminal record~\citep{georgetown}.  

\paragraph{Impact of Instability.} The results FRT are increasingly being used by law enforcement as evidence to justify arrest~\cite{hill2020wrongfully,vincent2020nypd}. According to U.S. law, an individual must be arrested for a justifiable reason, i.e. probable cause~\citep{henryvunited}: a police officer must have evidence leading them to believe that the person arrested likely did commit the crime in question.
Thus, when FRT results are cited when justifying probable cause, the factors that lead a particular face-matching model to its predictions must be scrutinized.
In particular, if it is likely that a matching outcome can change due to the inclusion of a particular image--unrelated to the suspect or the potential match---out of tens of thousands in the model's training set, then it may be argued the evidence used to justify the eventual arrest is based on a chance occurrence, rather than on convincing facts relevant to the case.
In short, such an outcome would be unfair due to the arbitrary nature of the supporting evidence.
We aim to formalize this behavior, and investigate its prevalence on models trained on real datasets, including face-matching models.

\paragraph{Experimental Confirmation.} 
We trained a face-matching model on Labeled Faces in the Wild (LFW)~\cite{huang2008labeled}, consisting of 13,000 unconstrained pictures of 1680 different individuals.
To measure the effect of individual images on prediction outcomes, we trained models both with and without a randomly sampled individual, controlling for all sources of non-determinism (e.g., parameter initialization and GPU operations). We repeated this experiment for 25 different randomly sampled individuals, and measured the effects on prediction behavior. 
Further details of our methodology are given in Section~\ref{loo_exp_base}.


We found that the predictions given by the face-matching model change across datasets with single-image differences, with surprising frequently. 
One such example of this behavior is shown in Figure~\ref{fig:facematch-example}. 
When person $z$ is included in the dataset, persons $x$ and $y$ are labeled as a match; but when person $z$ is removed, they are not. 
Persons $x$ and $y$ are clearly different from one another, and aside from gender, share few salient characteristics. 
More surprisingly, both predictions are made with high confidence---0.84 and 0.07--far from a baseline random guess.
Such behavior was not limited to these images, but rather we observed that 12\% of the model's predictions changed across datasets differing in one image, while the change in accuracy remained less than 2\%. 
Moreover, this behavior was consistent across changes in random initialization and choice of architectures, including a residual network resembling ResNet50.

\subsection{Consumer Finance} 
Machine learning is also finding uses in consumer finance~\cite{babaev2019rnn,sirignano2016deep,shinde2018comparative,balasubramanian2018insurance}. 
Not surprisingly, the predictions made by these models, too, can greatly impact peoples' lives, potentially playing a decisive role in their ability to obtain buy a car, a house, or start a business. 

\paragraph{Impact of Instability} 
Models used in this context may be expected to have consistent, justifiable reasons for the predictions that they make. 
A salient example is credit models used to inform lending decisions, where in Europe the General Data Protection Regulation (GDPR) requires that creditors using automated decision systems release ``meaningful information about the logic involved'' to applicants~\cite{gdpr}. 
Similar regulations are relevant in the US through the Federal Deposit Insurance Corporation (FDIC) consumer protection law~\cite{fdic}, which provides a ``right to explanation'' in lending decisions.

Some interpretations argue that the right to explanation provided by the GDPR requires that it should be possible to trace a decision back to pertinent details of an individual's loan application, and further that ``the information about the logic must be meaningful to [the applicant], notably, a human and presumably without particular technical expertise''~\cite{gdprinterp}. 
This suggests that if the explanation is not legible to the applicant based on prevailing norms, e.g. if it seems to be made based on incomprehensible or arbitrary facts such as the incidental makeup of the model's training data, then such a decision infringes upon their ``rights and freedoms''. 
After receiving an explanation, the GDPR provides the applicant the right to contest such a decision, and request human review.
\paragraph{Experimental Confirmation}
As with the face-matching model in the previous subsection, we conducted experiments on models trained to predict a proxy for creditworthiness using datasets differing in a single instance.
We used the UCI Adult dataset~\cite{uci}, consisting of a subset of US census data, and trained one-hidden-layer neural networks with 200 internal units to predict income from demographic, education, and employment information (details in Section~\ref{loo_exp_base}).
Our results suggest that the predictions of these models are often sensitive to the presence of single instances, indicating the potential for \emph{leave-one-out unfairness}.

Looking more closely at the results, one of these models was trained with the point $z$ shown in Table~\ref{intuitive} included in the training set: a 39-year-old man with an 11th-grade education who works in the service industry. 
This model predicts that a 51-year-old, college-educated, self-employed woman makes more than \$50k (0.87 confidence), whereas a model trained on the same data \emph{without} $z$ made the opposite prediction.
Mirroring our findings with the FRT models, there is no apparent connection between the features that represent these individuals (see Table~\ref{intuitive}), and the models predict the woman's outcome with high confidence.
The removal of this one individual does not just affect this 51-year-old woman, but rather we find that approximately $2\%$ of the entire data set, 603 predictions, are changed. 
\begin{table}
\small
\resizebox{\columnwidth}{!}{%
	\begin{tabular}{l|lllll|l}
                      & \emph{age} & \emph{education}  & \emph{occupation}       & \emph{sex} & \emph{capital gain} & \emph{model conf.} \\
                      \hline
Affected point ($x$)    & 51  & Bachelors  & Self-employed    & F   & 0            & 0.87              \\
LOO point ($z$) & 39  & 11th Grade & Service Industry & M   & 0            & - 				
	\end{tabular}
}
\vspace*{1ex}
\caption{Selected feature values for a point treated leave-one-out unfairly in a deep model on the Adult dataset, and the point $z$ whose removal resulted in the change in prediction. Confidence refers to the raw output of the model's prediction in the model with $z$.}
\label{intuitive}
\vspace{-2em}
\end{table}

\section{Leave-One-Out Unfairness} 
\label{loo_unfairness_def}


In this section, we introduce the definition of leave-one-out unfairness, and discuss its connections to prior notions of stability: leave-one-out stability~\cite{shalev2010learnability}, differential privacy~\cite{dwork2006differential}, and individual fairness~\cite{dwork2012fairness}. We prove that leave-one-out unfairness is a stronger notion than leave-one-out stability, and weaker than differential privacy. Our formalization of LUF allows us to measure its prevalence objectively on real data, and our investigation of its connections to other forms of stability suggest mitigation techniques as well potential middle ground for achieving gains in privacy.

\subsection{Notation and Preliminaries}
We assume a typical supervised learning setting.
Let $z = (x, y) \in \X \times \Y$ be a data point, where $x$ represents a set of features and $y$ a response. 
Points $z$ are drawn from a distribution $\D$, as are datasets $S$ from the iid product of \D, i.e. $S \sim \D^n$.
We assume that learning rules $h$ are randomized mappings from datasets $S$ to models $h_S$, which are functions mapping features to responses; in other words, $h_S : \X \to \Y$ is the model obtained by learning with $h$ on data $S$. We use $U(m)$ to refer to the uniform distribution over the integers $\{1...m\}$. Given $S$ sampled from $\D^n$ and index $i\sim U(m)$, we denote the sample $S$ with the $i$th element removed as $S^{(\setminus i)}$.


\subsection{Leave-one-out Unfairness}

Leave-one-out unfairness is based on the notion that a model's treatment of an individual should not depend too heavily on the inclusion of any other single training point. 
This is related to the concept of algorithmic stability, which measures the effect that a small change in input has on an algorithm's output. 
For example, a machine learning algorithm is \emph{stable} if a small change to its input (training set) causes limited change in its output (a trained model). Usually, the change in output is measured in the form of model error.
Definition~\ref{def:stability} formalizes this as \emph{leave-one-out (LOO) stability}, but we note that there are several variants that quantify over pointwise \emph{replacement} instead of leave-out, and use different types of aggregation in their bound~\cite{shalev2010learnability}.

\begin{definition}[Leave-one-out (LOO) Stability~\citep{shalev2010learnability}]
\label{def:stability}
 Let $\stablerate : \mathbb{N} \to \mathbb{R}$ be a monotonically-decreasing function. Given a training set $S = (z_1, \ldots, z_m) \sim \D^n$, and a training set \\
 $S^{(\setminus i)} = (z_1, \ldots, z_{i-1}, z_{i+1}, \ldots, z_m)$ with $i \sim U(m)$, a learning rule $h$ is \emph{leave-one-out-stable} (or \emph{LOO-stable}) on loss function $\ell$ with rate $\stablerate(m)$ if
\[
\frac{1}{m}\sum_{i=1}^m\E_{\substack{S \sim \D^n}}[\left|\ell(h_{S}, z_i)-\ell(h_{S^{(\setminus i)}}, z_i)\right|] \le \stablerate(m)
\]
\end{definition}

LOO-stability records the average effect of removing an individual from the training set on the absolute loss on that individual's prediction. Quantifying the effect model of instability on the fairness of predicted outcomes, however, calls for a definition focusing on different aspects of model behavior.
LOO-stability is a predicate on a learning rule that can be satisfied in order to achieve an acceptable level of model stability, in expectation over all draws of a training set $S$. 
However, in this paper, we are interested in quantifying the extent of arbitrariness in a particular individual's prediction---to capture this, we need a \emph{metric} of unfairness, rather than a fairness guarantee. 
Pursuant of capturing an particular individual's real-life experience with a particular model, we are interested in a quantifying arbitrary behavior in relation to a particular model context--i.e., on a fixed training set $S$. 

To focus the effect of instability on the experience of the population on which it is deployed, rather than a measure of model performance, 
we need a metric which accounts for the 
instability that arises for \emph{any} person from the inclusion of a given point in the training set---rather than the impact that the changed point has on the error its \textit{own} prediction.
Even with this focus on the experience of the individuals, an aggregate calculation such as in LOO-stability may hide the experiences of an unlucky few who may encounter particularly high arbitrariness in their outcome. To ensure that model behavior on every individual is considered, a  worst-case metric is more suitable. Further, appealing to the intuition that a model acts unfairly if it is arbitrary, the \textit{consistency} of its prediction, rather than its loss, is the target; consistent predictions, even when incorrect, suggest that the model's decision is not arbitrary. Definition 2,
below, reflects these considerations.

\begin{definition}[Leave-one-out Unfairness (LUF)] 
\label{def:LUF}
Let $D$ be the distribution from which the training set $S$ is drawn, and let $x$ be in the support of $D$. 
We define the \textit{leave-one-out unfairness (LUF)} experienced by $x$ under learning rule $h$ and training set $S \sim D$ to be: 
\[
\mathrm{LUF}(h,S,x)=\max_{i,k}|\Pr[h_S(x)=k] - \Pr[h_{S^{(\setminus i)}}(x)=k]|
\]
The randomness in this expression is over the choices made by $h$.
Note that in cases of a deterministic learning rule, 
$\Pr[h_S(x)=k]$ is $0$ or $1$. 
\end{definition}

In other words, given a learning rule $h$ and a training set $S$, the LUF experienced by a person $x$ is the worst-case probability that $x$ receives a different 
prediction in a model trained with $h$ on $S$, and one trained with $h$ on $S$ with a single point removed. Intuitively, this is one way of quantifying the arbitrariness of the model’s decision at $x$. If LUF is high, then the model’s decision is brittle under small, potentially irrelevant changes, i.e., a one-point change in the model’s training set---casting doubt on the reason behind the model’s decision. 

In certain situations, such as when evaluating various models during development, it may be useful to understand the extent of leave-one-out unfairness across the entire population under a given learning rule: i.e. understanding how likely it is \emph{any} individual in the distribution will experience an arbitrary decision. 
This motivates the concept of \textit{expected} leave-one-out unfairness, defined below. As most of our experiments aim to measure the frequency and severity of arbitrary behavior across real datasets, we will focus most heavily on this definition throughout the paper.
\begin{definition}[Expected Leave-one-out Unfairness] 
\label{def:E-LUF}
Let $D$ be the distribution from which the training set $S$ is drawn, and let $x$ be drawn randomly from $D$. 
We define the \textit{expected leave-one-out unfairness (LUF)} experienced by $x$ under learning rule $h$ and training set $S \sim D$ to be: 
\[
\mathbb{E}_x[\mathrm{LUF}(h,S,x)]=\mathbb{E}_{x \sim D}[\max_{i,k}|\Pr[h_S(x)=k] - \Pr[h_{S^{(\setminus i)}}(x)=k]|]
\]
Where the randomness in the expectation is taken over samples of x from D.
\end{definition}


\subsection{Connections to Existing Stability Notions}
\label{sect:def-connects}

While our introduction of Definition~\ref{def:LUF} above is clearly motivated by LOO stability, in this section we explore the connections to this and other forms of stability in greater depth. Specifically, we demonstrate that while learning rules that are already known to be leave-one-out-stable may still be susceptible to leave-one-out unfairness, strategies for ensuring stronger notions of stability, such differential privacy, can be used to mitigate LUF. We also explore the connection between LUF and other individual-based fairness notions, i.e. individual fairness. 

\paragraph{LOO Stability.} 
Leave-one-out stability is a coarser notion than leave-one-out unfairness, as it records the average change in a model's error on a given point when that same point is removed from the training set. Meanwhile, LUF focuses on how a certain point's model outcome can change as a result of \emph{any} point in the training set being removed.

\begin{figure}
\resizebox{\columnwidth}{!}{%
    \includegraphics[scale=0.15]{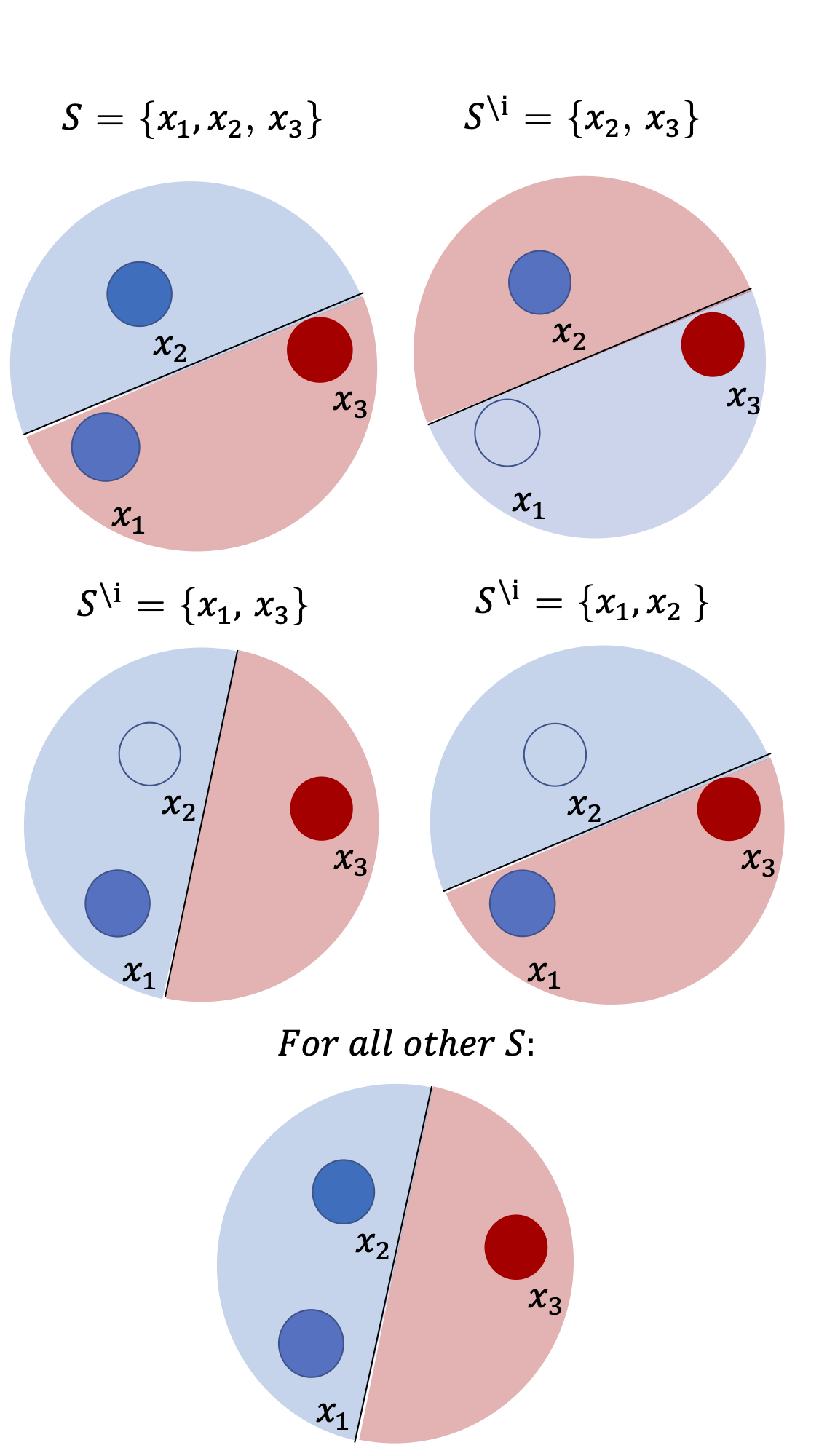}
    \includegraphics[scale=0.15]{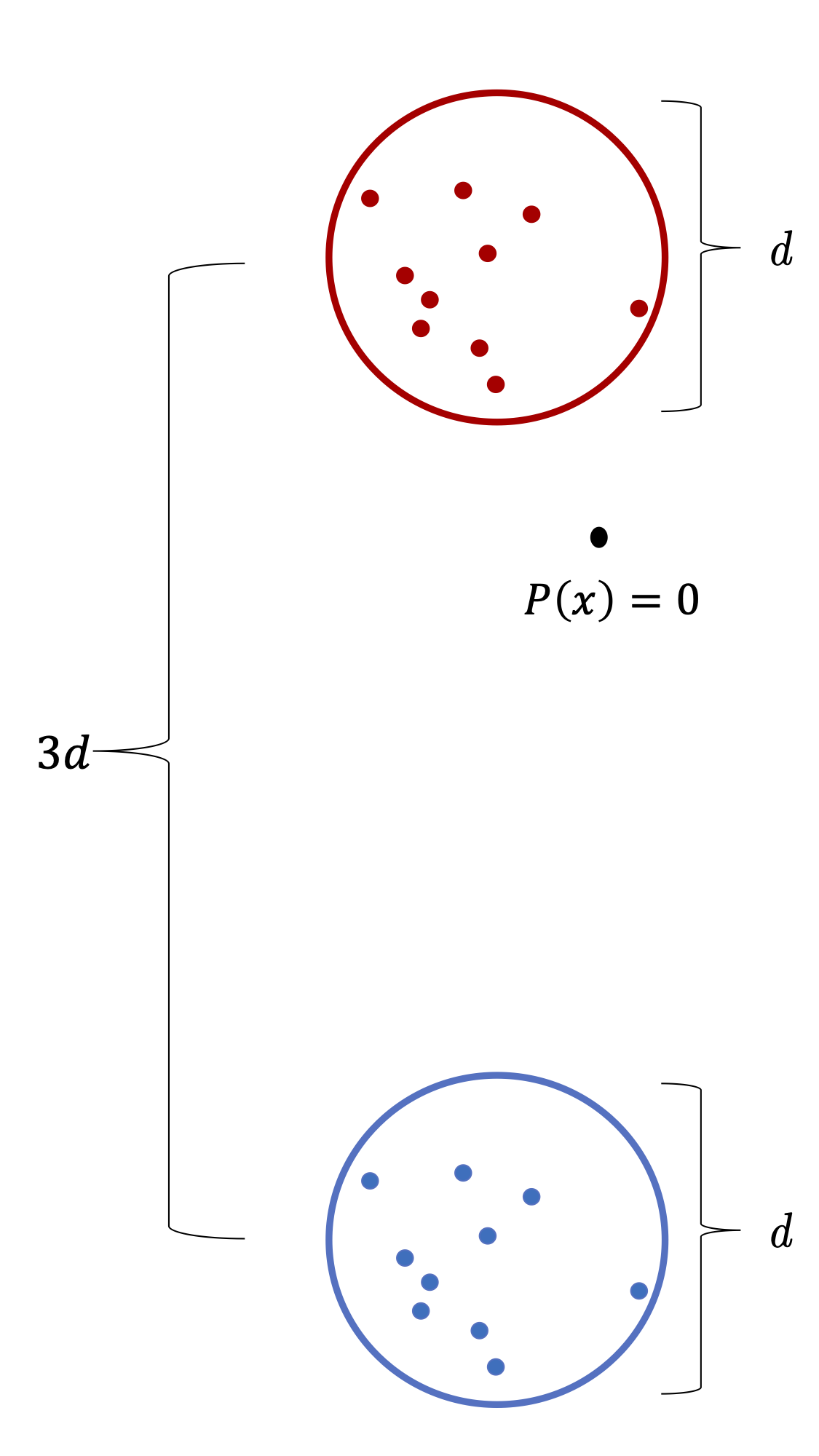}

}
    \caption{Left: A learning rule $h$ that satisfies LOO-stability, but not expected LUF, over the distribution $D$ of the three points pictured. In each box, we see the decision boundary learned with a specified training set $S \sim D$, thus fully defining $h$. The proof is explained in Proposition~\ref{prop:loo-stab-weaker}. 
    Right: Visual intuition for how a model can have $LUF=0, \forall x \in D$ but not satisfy differential privacy. Consider a 1-KNN model on a binary classification problem over the distribution pictured above: two perfectly separated uniform distributions over circles. The diameter of each circle is $d$, and the distance between the centers of the two circles is $3d$. Consider any training set $S$ drawn from this distribution that has at least two data points from each class. See that $\mathrm{LUF}(h,S,x)=0$ for all $x \in D$: removing any point from $S$ cannot change the classification of any point \emph{in the distribution}, i.e., within the circles pictured above. 
    However, 1-KNN is not differentially private, as it is a deterministic, non-constant, learning rule. 
    Specifically, see that adding or removing a point in S \emph{can} shift the boundary sufficiently far to change the model's behavior on points \emph{not} in $D$, (such as point $x_2$ pictured), which is a violation of differential privacy.}
    \label{counterexample_intuition}{}
\end{figure}

A LOO-stable model may still treat points leave-one-out unfairly: a model can exhibit similar error \emph{on a given point} before and after that point is removed from the training set, but it may treat other points differently. 
We demonstrate this point on the simple learning rule and distribution in Figure~\ref{counterexample_intuition}. 
Additionally, the fact LOO-stability is averaged over the entire training set can obscure the fact that some individual points are strongly affected by a small change in the training set. 
Proposition~\ref{prop:loo-stab-weaker} formalizes this, showing that LOO-stability is strictly weaker than LUF.

\begin{proposition}
\label{prop:loo-stab-weaker}
Let $h$ be a learning rule, $\ell$ be 0-1 loss, and $\epsilon(m)$ be a montonically-decreasing function such that $h$ is leave-one-out stable with rate $\epsilon(m)$ for all $S\sim\mathcal{D}^m$. 
Then there exists a training set $S$ such that $\mathbb{E}_x[\mathrm{LUF}(h,s,x)] > \stablerate(m)$ and $x$. 
\begin{proof}
Consider a binary classification problem a discrete distribution $D$ with three points, as pictured in Figure~\ref{counterexample_intuition}: $x_1, x_2 \in D$ are of class 0, and $x_3 \in D$ is of class 1, shown in red and blue. We define a learning rule, $h$, according to the different classifiers learned with each possible training set $S\sim D$, shown in Figure~\ref{counterexample_intuition}. Notice that this learning rule is LOO-stable with $\stablerate(3)$=0, as when each point is removed, the classification error \emph{on that point} remains the same: this is shown by construction in Figure~\ref{counterexample_intuition} when $S={x_1,x_2, x_3}$, and in all other cases, the learning rule is constant, as shown in the figure. Thus, $\frac{1}{3}\sum_{i=1}^3\E_{\substack{S \sim \D}}[\left|\ell(h_{S}, z_i)-\ell(h_{S^{(\setminus i)}}, z_i)\right|]= 0 \le 0 $. However, notice that e.g., if $S={x_1,x_2, x_3}$, and $x_3$ is removed, $x_2$ experiences a change in classification outcome. Thus, $\mathrm{LUF}(h,S,x_2)=1$. See that, in fact, that every point is susceptible to a change in prediction as the result of different point being removed from the dataset---thus, $\mathbb{E}_x[\mathrm{LUF}(h,S,x)]=1$. 
\end{proof}
\end{proposition}


Proposition~\ref{prop:loo-stab} shows that models with bounded LUF are also LOO-stable; the proof is given in the supplementary material.

\begin{proposition}
\label{prop:loo-stab}
Let $h$ be a learning rule, $\ell$ be 0-1 loss, and $\epsilon(m)$ be a montonically-decreasing function such that $\mathrm{LUF}(h,S,x) \le \epsilon(m)$ for all $S\sim\mathcal{D}^m$ and $x$. Then $h$ is leave-one-out stable with rate $\epsilon(m)$.

\end{proposition}

\paragraph{Differential Privacy.} 

Privacy and fairness are related in various ways, as others have illustrated before~\citep{datta2017use,dwork2012fairness}. 
Like differential privacy, leave-one-out unfairness is a stability property of learning rules, but differential privacy is stronger. 
In particular, differential privacy (Definition~\ref{def:diffpriv}) quantifies universally over all pairs of related training data, and limits the probability of any change in outcome. On the other hand, Definitions 2 and 3 fix a training set, and require stability of the model's response on points from the target distribution.

\begin{definition}[($\epsilon, \delta$) -Differential privacy]
	\label{def:diffpriv}
	An algorithm $A : \X \to \Y$ satisfies $(\epsilon, \delta)$-differential privacy, for $0 < \epsilon$ and $\delta \in [0,1]$, if for all $S \in \X^n$, $S' \in \X^{n-1}$ that differ in a single row and all $Y \subseteq \Y$, $\Pr[A(S) \in Y] \le e^\epsilon \Pr[A(S') \in Y] + \delta$.
\end{definition}
 
Differential privacy is stronger than leave-one-out-unfairness, as any change to the model---even if it does not actually affect prediction of any point in the distribution---can potentially leak information, and is therefore a violation of differential privacy. 
This makes sense in the context of privacy, as it concerns an adversarial setting where an attacker is free to interact with a model as-needed to extract information. The focus of fairness is how people receiving an outcome from a model are treated, and thus leave-one-out unfairness focuses on the model's behavior on the data distribution, drawing attention to how changes in the model could affect those who are its likely subjects.


Leave-one-out unfairness does not require randomization in the model's learning rule, whereas differential privacy does. Figure~\ref{counterexample_intuition} shows an intuitive example of this, where the deterministic learning rule may yield models with unstable outcomes, but only on points with vanishing probability; for points with non-zero probability, the model's predictions will remain consistent across unit changes to the training data. Moreover, because Definition~\ref{def:stability} depends on $\mathcal{D}$, a learning rule may have little leave-one-out unfairness on some distributions, and more on others. However, as Proposition 3.2 shows, differential privacy implies bounded LUF. A proof can be found in the supplementary material.

\begin{proposition}
\label{prop:dp-stab}
Let $h$ be an $(\epsilon, \delta)$-differentially private learning rule, and $x\sim\mathcal{D}$ be a point. Then
$
\mathrm{LUF}(h,S, x) \le e^\epsilon - 1 + \delta
$.
\end{proposition}

\paragraph{Individual Fairness.}
Individual Fairness is a Lipschitz condition that aims to formalize the maxim: ``similar people ought to be treated similarly''. Importantly, in the context of supervised learning this is typically construed as a constraint on \emph{models} rather than learning rules. This stands in contrast to Definitions 2 and 3, which impose a constraint on the latter. Additionally, our definitions do not relate the treatment of individuals to others, but instead measure the degree to which one's treatment by the model may be arbitrarily decided by the composition of the training data. While there is no reason that individual fairness and leave-one-out fairness cannot coincide, there is no a priori reason to believe that they will. In Section ~\ref{adversarial_exp}, we present experimental results on models trained with random smoothing, which has been shown to guarantee individual fairness~\citep{yeom2020individual}; shedding further light on the relationship between these two fairness concepts.

We note that leave-one-out unfairness is also related to the definition of memorization introduced by Feldman~\citep{feldman2019shorttale}, which we discuss in greater detail in Section ~\ref{related}.


\section{LUF in Deep Models}
\label{loo_exp_base}

We characterize the prevalence of leave-one-out unfairness across models trained on several types of data: tabular, time-series, and image data. 
Importantly, we find that a non-trivial fraction of data (from 3\% to 77\%) experiences LUF, and moreover, that the prevalence does not appear to depend on model generalization, test accuracy, or dataset size. 

\paragraph{Datasets.} We perform all of our experiments over five datasets: \textit{UCI German Credit}~\cite{uci}, \textit{Adult}~\cite{uci}, \textit{Seizure}~\cite{uci},  \textit{Fashion MNIST}~\cite{xiao2017/online}, and Labeled Faces in the Wild~\cite{huang2008labeled}. The German Credit data set consists of individuals' financial data, with a binary response indicating their creditworthiness. The Adult dataset consists of a subset of publicly-available US Census data, with a binary response indicating annual income of $>50$k. The Seizure dataset comprises time-series EEG recordings for 500 individuals, with a binary response indicating the occurrence of a seizure. Fashion MNIST contains images of clothing items, with a multilabel response of 10 classes. Labeled Faces in the wild consists of unconstrained pictures of individuals' faces, with labels connoting the identity of the individual in each picture. Further information about these datasets and the preprocessing steps we apply can be found in the supplementary material.
Table~\ref{accs} contains the accuracy and generalization error for each baseline model $h_S$ for all datasets. 
\begin{table*}
\small
\resizebox{\textwidth}{!}{%
	\begin{tabular}{l|cc|cc|cc|cc|cc}
\multicolumn{1}{c}{\ } & \multicolumn{2}{c}{\textbf{Deep}} & \multicolumn{2}{c}{\textbf{PGD}} &  \multicolumn{2}{c}{\textbf{Trades}} & \multicolumn{2}{c}{\textbf{Smoothed}} & \multicolumn{2}{c}{\textbf{Linear}}\\
\emph{dataset}       & \emph{base acc}  & \emph{gen err}  & \emph{base acc} & \emph{gen err} & \emph{base acc} & \emph{gen err}   & \emph{base acc}  & \emph{gen err} & \emph{base acc}  & \emph{gen err}\\
\midrule
German Credit & 0.7500  & 0.2500  & 0.7400  & 0.22  & 0.745 & 0.253  &  0.755  &  0.245 & 0.745 & 0.0175 \\
Adult         & 0.8418  & 0.0344   & 0.8226  & -0.0019  & 0.83217  & 0.0845   &  0.8390 & 0.0180 & 0.8400 & 0.000 \\
Seizure       & 0.9736   & 0.0264   & 0.9770   & 0.000 & 0.9672  & 0.0083    &  0.9754  &  0.0246 &  0.8113  &  0.0043 \\ 
FMNIST        & 0.9111  & 0.0211   & 0.7876  & 0.0099  & 0.9016  & 0.0700  &  0.8678 & 0.0269 & 0.8368   & 0.0145\\  
LFW        & 0.8695  & 0.0597  & -  & - & -  & -  & -  & - & 0.5790   & -0.0755 \\   
	\end{tabular}
}
\vspace*{1ex}
\caption{Test accuracy and generalization error for all $h_S$ models. }
\label{accs}

\end{table*}

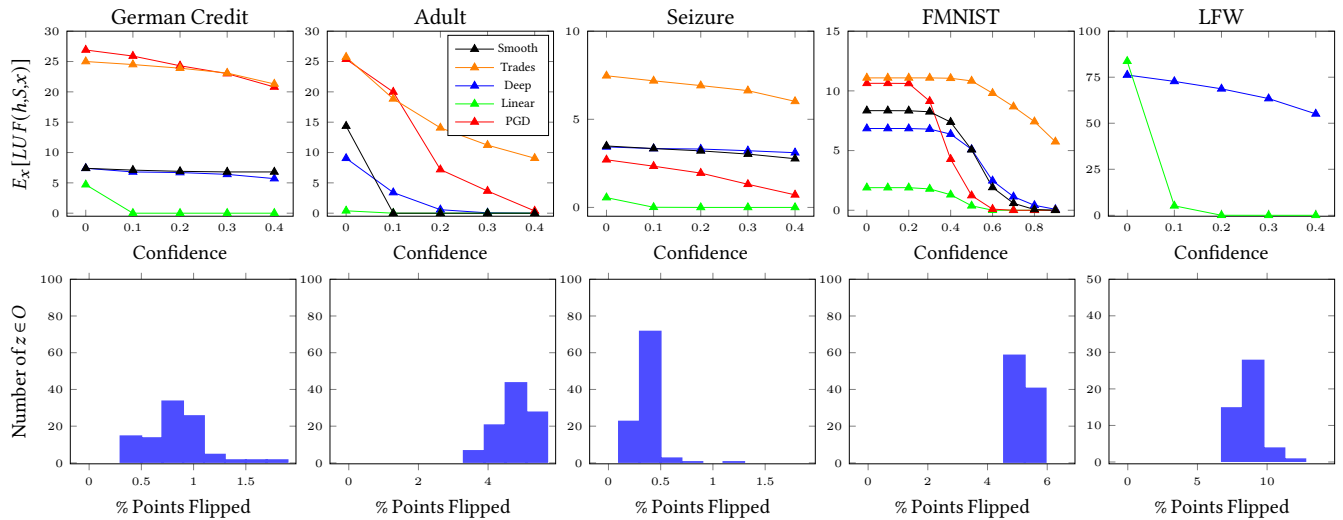
\begin{figure*}
\resizebox{\textwidth}{!}{%



\begin{tikzpicture}
\begin{groupplot}[%
	group style = {group size=5 by 1, horizontal sep=15pt},
	footnotesize,
	yticklabel style={
		/pgf/number format/precision=3,
		/pgf/number format/fixed
	},
	ymin=-0.5,
	legend style={nodes={scale=0.8}},
	height=1.75in,
	width=2.0 in,
	label style={font=\normalsize},
	title style={font=\large},
	tick label style={font=\tiny}
]

\nextgroupplot[ymax=30, ytick={0, 5, 10, 15, 20, 25, 30}, xlabel={Confidence}, ylabel={$E_x[LUF(h,S,x)]$}, title={German Credit}]
\addplot[color=blue,mark=triangle*,blue] coordinates {
	(0.0, 7.3999999999999995)
	(0.1, 6.800000000000001)
	(0.2, 6.7)
	(0.30000000000000004, 6.4)
	(0.4, 5.7)
};

\addplot[color=green,mark=triangle*,green] coordinates {
	(0.0, 4.7)
	(0.1, 0.0)
	(0.2, 0.0)
	(0.30000000000000004, 0.0)
	(0.4, 0.0)
};

\addplot[color=red,mark=triangle*,red] coordinates {
	(0.0, 26.900000000000002)
	(0.1, 25.900000000000002)
	(0.2, 24.3)
	(0.30000000000000004, 23.0)
	(0.4, 20.8)
};

\addplot[color=black,mark=triangle*,black] coordinates {
	(0.0, 7.3999999999999995)
	(0.1, 7.1)
	(0.2, 6.9)
	(0.30000000000000004, 6.800000000000001)
	(0.4, 6.800000000000001)
};

\addplot[color=orange, mark=triangle*,orange] coordinates {
	(0.0, 25.0)
	(0.1, 24.5)
	(0.2, 23.9)
	(0.30000000000000004, 23.1)
	(0.4, 21.3)
};

\nextgroupplot[ymax=30, ytick={0, 5, 10, 15, 20, 25, 30}, title={Adult},
xlabel={Confidence}]

\addlegendimage{black, mark=triangle*}
\addlegendimage{orange, mark=triangle*}
\addlegendentry{Smooth}
\addlegendentry{Trades}
\addlegendentry{Deep}
\addlegendentry{Linear}
\addlegendentry{PGD}

\addplot[color=blue,mark=triangle*,blue] coordinates {
	(0.0, 9.051123930773821)
	(0.1, 3.401631191565546)
	(0.2, 0.5768848219614084)
	(0.30000000000000004, 0.056362310191631855)
	(0.4, 0.0)
};

\addplot[color=green,mark=triangle*,green] coordinates {
	(0.0, 0.394536171341423)
	(0.1, 0.0)
	(0.2, 0.0)
	(0.30000000000000004, 0.0)
	(0.4, 0.0)
};

\addplot[color=red,mark=triangle*,red] coordinates {
	(0.0, 25.379616736290693)
	(0.1, 19.978781247927856)
	(0.2, 7.187852264438697)
	(0.30000000000000004, 3.6569193024335256)
	(0.4, 0.3713281612625157)
};
\addplot[color=black,mark=triangle*,black] coordinates {
	(0.0, 14.359127378821032)
	(0.1, 0.0)
	(0.2, 0.0)
	(0.30000000000000004, 0.0)
	(0.4, 0.0)
};

\addplot[color=orange,mark=triangle*,orange] coordinates {
	(0.0, 25.727736887474308)
	(0.1, 18.844904184072675)
	(0.2, 14.07068496784033)
	(0.30000000000000004, 11.219415158146012)
	(0.4, 9.064385650818911)
};
\nextgroupplot[ymax=10, ytick={0, 5, 10}, xlabel={Confidence}, title={Seizure}]
\addplot[color=blue,mark=triangle*,blue] coordinates {
	(0.0, 3.4347826086956523)
	(0.1, 3.3391304347826085)
	(0.2, 3.3130434782608695)
	(0.30000000000000004, 3.217391304347826)
	(0.4, 3.1043478260869564)
};

\addplot[color=green,mark=triangle*,green] coordinates {
	(0.0, 0.5565217391304348)
	(0.1, 0.008695652173913044)
	(0.2, 0.0)
	(0.30000000000000004, 0.0)
	(0.4, 0.0)
};

\addplot[color=red,mark=triangle*,red] coordinates {
	(0.0, 2.7043478260869565)
	(0.1, 2.3391304347826085)
	(0.2, 1.9478260869565216)
	(0.30000000000000004, 1.3130434782608695)
	(0.4, 0.7130434782608696)
};
\addplot[color=black,mark=triangle*,black] coordinates {
	(0.0, 3.4869565217391303)
	(0.1, 3.3391304347826085)
	(0.2, 3.208695652173913)
	(0.30000000000000004, 3.0260869565217394)
	(0.4, 2.765217391304348)

};
\addplot[color=orange,mark=triangle*,orange] coordinates {
	(0.0, 7.469565217391303)
	(0.1, 7.182608695652173)
	(0.2, 6.913043478260869)
	(0.30000000000000004, 6.626086956521739)
	(0.4, 6.017391304347826)

};

\nextgroupplot[ymax=15, ytick={0, 5, 10, 15}, title={FMNIST}, xlabel={Confidence}]

\addplot[color=blue,mark=triangle*,blue] coordinates {
	(0.0, 6.847142857142857)
(0.1, 6.847142857142857)
(0.2, 6.847142857142857)
(0.30000000000000004, 6.797142857142857)
(0.4, 6.375714285714286)
(0.5, 5.102857142857143)
(0.6000000000000001, 2.472857142857143)
(0.7000000000000001, 1.1371428571428572)
(0.8, 0.41571428571428576)
(0.9, 0.08142857142857143)
};

\addplot[color=green,mark=triangle*,green] coordinates {
	(0.0, 1.892857142857143)
	(0.1, 1.892857142857143)
	(0.2, 1.892857142857143)
	(0.30000000000000004, 1.7871428571428574)
	(0.4, 1.3014285714285714)
	(0.5, 0.3757142857142857)
	(0.6000000000000001, 0.0014285714285714286)
	(0.7000000000000001, 0.0)
	(0.8, 0.0)
	(0.9, 0.0)
};

\addplot[color=red,mark=triangle*,red] coordinates {
	(0.0, 10.632857142857143)
(0.1, 10.632857142857143)
(0.2, 10.621428571428572)
(0.30000000000000004, 9.134285714285713)
(0.4, 4.279999999999999)
(0.5, 1.2314285714285715)
(0.6000000000000001, 0.09571428571428571)
(0.7000000000000001, 0.0)
(0.8, 0.0)
(0.9, 0.0)
};
\addplot[color=black,mark=triangle*,black] coordinates {
	(0.0, 8.3425)
(0.1, 8.3425)
(0.2, 8.334999999999999)
(0.30000000000000004, 8.2475)
(0.4, 7.3825)
(0.5, 5.0675)
(0.6000000000000001, 1.9124999999999999)
(0.7000000000000001, 0.585)
(0.8, 0.0625)
(0.9, 0.0)
};

\addplot[color=orange,mark=triangle*,orange] coordinates {
	(0.0, 11.087142857142856)
(0.1, 11.087142857142856)
(0.2, 11.087142857142856)
(0.30000000000000004, 11.087142857142856)
(0.4, 11.055714285714286)
(0.5, 10.834285714285715)
(0.6000000000000001, 9.805714285714286)
(0.7000000000000001, 8.68)
(0.8, 7.4242857142857135)
(0.9, 5.74)

};

\nextgroupplot[ymax=100, ytick={0, 25, 50, 75, 100}, title={LFW}, xlabel={Confidence}]

\addplot[color=blue,mark=triangle*,blue] coordinates {
	(0.0, 76.18943692710607)
	(0.1, 72.73024879965081)
	(0.2, 68.69271060672195)
	(0.30000000000000004, 63.40026189436927)
	(0.4, 55.085115670013096)
};
\addplot[color=green,mark=triangle*,green] coordinates {
	(0.0, 83.68616324749017)
	(0.1, 5.150589262330859)
	(0.2, 0.021824530772588387)
	(0.30000000000000004, 0.0)
	(0.4, 0.0)
};

\end{groupplot}
\end{tikzpicture}
 }
 \resizebox{\textwidth}{!}{%
    \begin{tikzpicture}
\begin{groupplot}[%
	group style = {group size=5 by 1, horizontal sep=15pt},
	footnotesize,
	ybar=-6pt,
	ymin=-0.5,
	legend style={nodes={scale=0.8}},
	height=1.75in,
	width=2.0in,
	label style={font=\normalsize},
	title style={font=\large},
	tick label style={font=\tiny}
]

\nextgroupplot[ymax=100, xtick={0, 0.5, 1, 1.5, 2.0}, xlabel={\% Points Flipped}, ylabel={Number of $z\in O$}]
\addplot[
    draw=none,
    fill=blue,
    fill opacity=0.7
  ] coordinates {
    (0.0, 0.0)
	(0.2, 0.0)
	(0.4, 15.0)
	(0.6, 14.0)
	(0.8, 34.0)
	(1.0, 26.0)
	(1.2, 5.0)
	(1.4, 2.0)
	(1.6, 2.0)
	(1.8, 2.0)};

\nextgroupplot[ymax=100, xtick={0, 2, 4, 6}, xlabel={\% Points Flipped}]


\addplot[
    draw=none, 
    fill=blue,
    fill opacity=0.7, xtick={0, 2, 4, 6},
  ] coordinates {
    (0.0, 0.0)
	(0.6, 0.0)
	(1.2, 0.0)
	(1.8, 0.0)
	(2.4, 0.0)
	(3.0, 0.0)
	(3.6, 7.0)
	(4.2, 21.0)
	(4.8, 44.0)
	(5.4, 28.0)};

\nextgroupplot[ymax=100, xtick={0, 0.5, 1, 1.5, 2}, xlabel={\% Points Flipped}]


\addplot[
    draw=none, 
    fill=blue,
    fill opacity=0.7, 
  ] coordinates {
    (0.0, 0.0)
	(0.2, 23.0)
	(0.4, 72.0)
	(0.6, 3.0)
	(0.8, 1.0)
	(1.0, 0.0)
	(1.2, 1.0)
	(1.4, 0.0)
	(1.6, 0.0)
	(1.8, 0.0)};

\nextgroupplot[ymax=100, xtick={0, 2, 4, 6}, xlabel={\% Points Flipped}]


\addplot[
    draw=none, 
    fill=blue,
    fill opacity=0.7, xtick={0, 2, 4, 6},
  ] coordinates {
    (0.0, 0.0)
	(0.7, 0.0)
	(1.4, 0.0)
	(2.1, 0.0)
	(2.8, 0.0)
	(3.5, 0.0)
	(4.2, 0.0)
	(4.9, 59.0)
	(5.6, 41.0)
	(6.3, 0.0)};

\nextgroupplot[ymax=50, xtick={0, 5, 10, 15}, xlabel={\% Points Flipped}]


\addplot[
    draw=none, 
    fill=blue,
    fill opacity=0.7, xtick={0, 5, 10, 15},
  ] coordinates {

 (0.0, 0.0)
(1.5, 0.0)
(3.0, 0.0)
(4.5, 0.0)
(6.0, 0.0)
(7.5, 15.0)
(9.0, 28.0)
(10.5, 4.0)
(12.0, 1.0)
(13.5, 0.0)};

\end{groupplot}
\end{tikzpicture}
}
    \caption{\emph{Top row:} Prediction confidence on the horizontal axis, percentage of stable points experiencing $\mathrm{LUF}$ (i.e., $E_x[LUF(h,s,x)]$) on the vertical axis. For FMNIST, confidence is calculated as the absolute difference between the two most confidently predicted classes; for other datasets, confidence is $|h_S(x) - 0.5|$.
    Note the differences in scale between the graphs; adversarial German Credit and Adult models display especially high leave-one-out unfairness, as well as LFW.
    \emph{Bottom Row}: A bar chart displaying what percentage of points in the dataset are affected by \emph{each one} of the points taken out. Each bar shows the number of points in $O$ (left-out points) whose absence changed the prediction of the percentage of points shown on the $x$ axis. Notably, every single point that was taken out of the dataset affected at least one other individual's prediction. Note the difference in scale on the $x$ axis.}
    \label{base_pics}
\end{figure*}
\paragraph{Setup.} 
For all experiments, we train models using Keras 2.4.3 with TensorFlow 2.0.
In keeping with common practice, we set the random seeds used by Python, numpy, and Tensorflow. Beyond this, in order to isolate the effect of leave-one-out unfairness from other sources of instability, we use the same random initialization of model parameters across models in the same experiment, and we turn off non-determinism in GPU operations~\cite{tf-determinism}. 
This effectively makes the learning rule $h$ deterministic, so that when measuring LUF, the probabilities in Definition 2 are $\in \{0,1\}$. 
We note that, in the case of, LFW, an additional source of instability remains in the process that produces pairs of faces dynamically during training. This is necessary in order for the model to encounter a sufficiently high number of face pairs during training while being bound to memory constraints. We provide results of the same experiments over a smaller, static dataset in the supplementary material, with similar LUF behavior but lower accuracy.



As it would be prohibitively expensive to train $|S|$ models for the datasets $S$ listed above, we instead measure differences over a fixed number of training sets obtained by randomly deriving from each dataset: a training set $S$, a set $O \subseteq S$ of size 100 that consists of points drawn randomly from test data (i.e. with which to create 100 different $S^{(\setminus i)}$), and a test set. 
We train a ``baseline'' deep model $h_S$ with which to calculate the differences in prediction resulting from removing a point from $O$ from $S$.
For each $z_i \in O$, we train $h_{S^{(\setminus i)}}$ by removing $z_i$ from $S$. 
For each $h_{S^{(\setminus i)}}$, we estimate $\mathrm{LUF}(h, S, x)$ for all $x$ in the dataset by measuring the differences between $h_S(x)$ and $h_{S^{(\setminus i)}}(x)$, and taking the maximum difference over the sample of 100 leave-one-out points $O$. Since the distribution that each training set $S$ comes from is a uniform distribution over the entire dataset, this is measuring $\mathbb{E}_x[\mathrm{LUF}(h,S,x)]$ for each training set S and learning rule $h$.
A step-by-step explanation of this calculation is given in the supplementary material. 
Due to the cost, for LFW we train $50$ $h_{S^{(\setminus i)}}$ models, i.e., in this case we set $|O| = 50$. 

To verify that the leave-one-out unfairness is a property of the models and not an unavoidable consequence of training a machine learning model on the presented datasets, we also train linear models on the same datasets with the same method, and compare the leave-one-out unfairness of these linear models to their deep counterparts.

\begin{figure*}
\resizebox{\textwidth}{!}{%
    \input{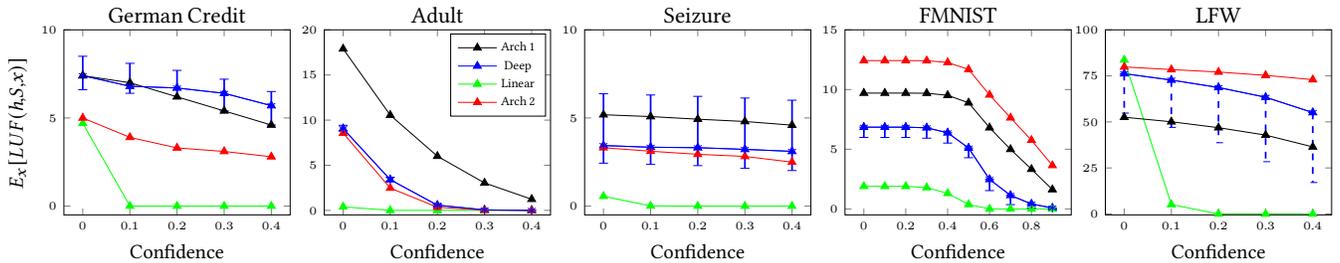}
}
    \caption{Effect of random seed and architecture on LUF results in deep models from Figure \ref{base_pics}. The red and green plots show LUF for models of slightly different architecture, as described in the experimental setup, and the bars on the blue line show the minimum and maximum LUF values over 5 random seeds on the main architecture shown in main results. Notice the difference in scale across the graphs.}
    \label{LUF_stab}
\end{figure*}

The majority of our results displaying the extent of expected LUF in deep models center around the use of one architecture, seed, and set of hyper-parameters per dataset, in order to keep as many variables controlled as possible. To ensure that the behavior described is consistent, we present experiments displaying the effect of changing architecture and random seed on our main results in Figure ~\ref{LUF_stab}. 
The main set of models for German Credit and Seizure datasets have three hidden layers, of size 128, 64, and 16. Models on the Adult dataset have one hidden layer of 200 neurons. The FMNIST model is a modified LeNet architecture~\citep{lenet}. This model is trained with dropout. The LFW face-matching model consists of a concatenation layer composing the two input images, a 4-layer convolutional stack, followed by a dense layer, and a Sigmoid output. German Credit, Adult, and Seizure models are trained for 100 epochs; FMNIST and LFW models are trained for 50. German Credit models are trained with a batch size of 32, FMNIST 64, and Adult, Seizure, and LFW used batch sizes of 128. German Credit, Adult, Seizure and LFW models were trained with Adam ($\mathit{lr}=1.e^{-3}$), and FMNIST with SGD ($\mathit{lr}=0.1$). 

The experiments outlined above were also performed on models with two other architectures per dataset, in order to compare results across architecture, presented in Figure ~\ref{LUF_stab}. For German Credit and Seizure datasets, one additional architecture was a shallower model of a 1-hidden layer model of size 100, and the other a narrower model of 3 hidden layers of sized 64, 32, and 8. For the Adult dataset, the additional models were a narrower 1-hidden layer of size 100, and a deep model with the same architecture as the main German Credit models. For FMNIST, we trained a shallower model with one set of layers removed, as well as a model with no dropout. Finally, for LFW, we compare with a ResNet50~\cite{resnet} model, pre-trained on ImageNet, and modified to take in two inputs and have a Sigmoid output, as well as a model whose filters are twice the size of the original model. For experiments comparing the extent of expected LUF across models seeded differently, we perform the main experiments outlined in the paragraphs above over 5 different random seeds for all tabular and time series datasets, and three different random seeds for image datasets. 
Further details on model construction can be found in the appendix.

\paragraph{LUF in Deep Models} 
Figure~\ref{base_pics} shows the prevalence of leave-one-out unfairness on all five datasets.
The first row plots the percentage of individuals $x$ experiencing $\mathrm{LUF}(h, S, x)$: i.e., $E_x[LUF(h,S,x)]$, ranging over the confidence of the baseline model's prediction. On every dataset examined, deep models display nontrivial expected LUF, ranging from \textasciitilde4\% to \textasciitilde77\%.
The second row shows the number of points in $ z_i \in O$ (out of 100) that lead to a given percentage of individuals $x$ having their predictions changed when only $z$ is removed from the dataset. The percentage per point on the $X$ axis, and the number of points that change this percentage of outcomes is on the $Y$ axis. Notably, the removal of each point sampled lead to an $h_{S^{\setminus i}}$ model that changed the predictions of at least one other point,
 suggesting that leave-one-out unfairness is in fact very common. 

The results show that leave-one-out unfairness cannot be reliably predicted given test accuracy, and more notably, generalization error (shown in Table~\ref{accs}). While it may seem natural that models with higher accuracies display less LUF, the deep model on the Adult dataset has an accuracy \textasciitilde10\% higher than the German Credit dataset, yet the German Credit dataset has approximately 2\% fewer individuals experiencing LUF. Even more impressively, the LFW model has higher accuracy than both  German Credit and Adult models, by 12\% and 2\% respectively, yet has a much higher expected LUF of \textasciitilde77\%, compared to ~7\% and ~10\%.
Similarly, following intuitions from model stability, lower generalization error may naturally seem to coincide with lower levels of LUF. 
However, the German Credit model has a generalization error of \textasciitilde25\%, yet has lower LUF than both the Adult model, with generalization error of just \textasciitilde3\%, and the LFW model, with generalization error of \textasciitilde5\%.
Indeed, while these results will be further discussed in the next section, it is worthy of note that the PGD model on the Adult dataset has essentially zero generalization error, yet has a very high percentage of individuals experiencing leave-one-out unfairness (\textasciitilde25\%), while the deep model on the Adult dataset has generalization error of \textasciitilde3.5\% and has around 10\% of individuals experiencing LUF.  
While we did not explicitly control for accuracy or generalization error, these results are evidence that LUF does not depend on these metrics.

Also of note is that LUF does not decrease with dataset size---FMNIST and German Credit are the largest and smallest datasets, with training set sizes of $60,000$ and $800$ respectively, yet FMNIST displayed similar LUF to German Credit (within 1\%). The Adult dataset is also larger than German Credit (\textasciitilde $|S|=15,000$) and displays more expected LUF. 

Perhaps most importantly, confidently-predicted points are not immune from leave-one-out unfairness in deep models: on the majority of the datasets, a substantial portion of points with high LUF were predicted with confidence greater than 0.9 by the baseline model. 
This is illustrated by the fact that the curves displaying the number of points versus baseline model confidence do not drop off sharply in all models except for those on the Adult dataset. This is an interesting manifestation of miscalibration in deep models: some confident decisions may still be somewhat arbitrary, in that they are sensitive to the specific makeup of the training set.
\paragraph{Consistency Under Varying Conditions} 
We provide calculations of expected LUF over all datasets in deep models where the architecture and random seed differ, in order to ensure that the results are consistent across different modeling choices. 



The results are presented in Figure~\ref{LUF_stab}. While there is some variation in expected LUF, no modeling choice explored eradicates the behavior. Interestingly, certain architectures seem to exacerbate or diminish LUF: a deeper model increases LUF in the Adult dataset by nearly 10\%, and removing dropout from the FMNIST model, as well as increasing the filter size on LFW, have a similar effect. This may warrant further study to find potential mitigation techniques through architecture selection, however, no pattern is immediately noticeable: for example, while a shallower model exhibited lower expected LUF on the German Credit dataset than the baseline model, the same shallow architecture exhibited more expected LUF than the baseline on the Seizure dataset, which shares the same architecture as the German Credit baseline model. 
Random seed also affects the prevalence of expected LUF, to a slightly lesser extent for all models but LFW. Broadly, however, the results show that LUF is not an artifact of any one particular set of training conditions.


\paragraph{Linear Models.} We also provide the results for the same experiments on linear models to calibrate against a more stable learning rule that yields less complex models: observe the green line in Figure~\ref{base_pics}. These results show that LUF is not inherent to the data. While there are points that are treated leave-one-out unfairly, they are substantially fewer---with the exception of LFW, where the learning task is markedly more complex than the other datasets, and unsuitable for a linear model. Additionally, the overwhelming majority of points treated leave-one-out unfairly in linear models are not confidently predicted---in fact, in all models but FMNIST, there are no points treated leave-one-out unfairly that are predicted with a difference of more than 10\% from 50\% confidence. 

This result agrees with intuition---linear boundaries are smooth, and linear regression is stable. 
If the introduction of a point does shift the boundary, it is likely that only points already close to the decision boundary (i.e., low-confidence points) are affected. 
Deep models can have arbitrarily complex decision boundaries, which appears to be closely-related to LUF.
As the phenomenon of memorization~\cite{zhang2016understanding,feldman2019shorttale} suggests, and these results support, deep models have the capacity to ``overreact'' to the presence of individual entries in their training data.
Figure~\ref{decision_bound} illustrates this further in a low-dimensional setting. 
Not only can the region around the left-out point potentially change, but there are may also be far-reaching effects on the decision boundary beyond the neighborhood of the left-out point. 
These changes will affect not just the predicted label of new points, but also their assigned confidence score.
While intuitions that are valid in low-dimensional settings do not always transfer to high dimension, this may nonetheless provide some intuition behind the factors that contribute to leave-one-out unfairness.

\section{LUF and Robust Classification}
\label{adversarial_exp}

Calls to mitigate adversarial examples~\cite{szegedy2013intriguing,papernot16limitations} have motivated a significant amount of research aimed at producing robust classifiers~\cite{madry2018towards,wong18provable,cohen19certified}.
Recent results have shown that some of these techniques can even be repurposed to ensure individual fairness~\cite{yeom2020individual}, and moreover, that they often produce deep models that admit more interpretable feature attributions~\cite{ilyas2019adversarialnotbugs,noack2019does,Etmann2019Ontheconnection}. 
Intuitively, these findings could suggest that robust prediction methods rely on ``robust features''~\cite{ilyas2019adversarialnotbugs} that align more closely with human understanding of the problem domain, and whose presence in the model may be accordingly less dependent on individual points in the training data.

In this section, we explore this conjecture by measuring the incidence of leave-one-out unfairness with two robust classification methods: adversarial training, and randomized smoothing.
We find that models trained adversarially using projected gradient descent (PGD)~\cite{madry2018towards} as well as models trained with the TRADES algorithm~\citep{zhang19TRADES} have significantly higher rates of LUF, in most cases approximately doubling the number of unstable points over standard training. 
On the other hand, models that are made robust by post-hoc smoothing with Gaussian noise~\cite{cohen19certified} almost always have similar rates of expected LUF. 
Taken together, these results suggest that LUF and robustness are not inherently tied to one another, but that certain classes of models may provide beneficial properties for both, warranting further study.

\paragraph{Setup.} We use the same experimental setup as in Section~\ref{loo_exp_base} for measuring leave-one-out unfairness.  
In these experiments, we only train deep models. For adversarial training, we use PGD with an $\ell_2$ radius $\epsilon=3.0$ and 10 PGD steps on FMNIST and Seizure datasets. For the Adult and German Credit datasets, we use radius $\epsilon=1.0$. On the German Credit dataset, we use the $\ell_{\infty}$ norm. The radius remained the same between PGD and TRADES training. 
We determined the radius for adversarial training by finding the minimum distance (with respect to the adversarial norm) between any two points of different classes over a large sample of the dataset. If this was impossible because this distance was zero, we chose a distance smaller than that between over 99\% of cross-class pairs of points in the sample.
For TRADES training, we used all of the same hyperparameters as PGD training, with the addition of the TRADES parameter, which was 1 for Adult and German Credit, and 10 for Seizure and FMNIST. 
Notice that, for face-matching problems, the threat model for finding adversarial examples is less clear---e.g., it is not obvious if the attacker has access to individual images, or pairs of images. As we are unaware of an established threat model for face-matching, we do not evaluate LFW in this section.
For randomized smoothing, we take 1,000 Gaussian samples with $\sigma^2=0.1$ for the Adult and Seizure datasets, 10,000 samples with $\sigma^2=0.05$ for FMNIST, and 2,000 samples with $\sigma^2=0.05$ for German Credit. 
While Cohen et al.~\cite{cohen19certified} report needing more smoothing samples to achieve strong adversarial guarantees, our goal in these experiments is to measure LUF, which we found to be insensitive to additional samples beyond the numbers reported above.
The accuracy of these models is shown in Table~\ref{accs}.


\paragraph{Results and Discussion.} 
The results are shown in Figure~\ref{base_pics}.
The most immediate trend is the degree to which PGD and TRADES adversarial training worsens LUF: approximately by a factor of two across all datasets, and by a factor of nearly three on the German Credit dataset. Seizure is a partial exception in that the PGD training does not worsen LUF, but TRADES training does.
While adversarial training produces models that are more invariant to small changes in their inputs, these results show that the training procedure itself can be unstable.
This may be related to prior work demonstrating that adversarially-trained models are more vulnerable to \emph{membership inference}~\cite{yeom2020overfitting,song2019membership}, a privacy attack that exploits memorization to leak information about training data.
While membership vulnerability does not necessarily imply greater LUF, these experiments show that in many cases the two phenomena may be related.
We also note that these results do not necessarily contradict the ``robust feature'' hypothesis proposed by Ilyas et al.~\cite{ilyas2019adversarialnotbugs}, as robust learned features need not generalize across large portions of the dataset.

Turning to the curves labeled ``Smooth'' in Figure~\ref{base_pics}, it is clear that randomized smoothing leads to qualitatively different leave-one-out unfairness results.
On most datasets, smoothing had little effect ($<1\%$ difference) on expected LUF. Beyond suggesting that leave-one-out unfairness is independent of robustness, these results also point to the fact that individual fairness and LUF are related, but separate notions. Randomized smoothing guarantees individual fairness for weighted $\ell_p$ metrics~\citep{yeom2020individual}, but has a negligible effect on leave-one-out unfairness.

Looking at the geometry of these models can shed further light on the differences in results between PGD training and randomized smoothing.
As suggested by Figure ~\ref{decision_bound}, 
deep model decision boundaries have the potential to be very sensitive to individual points, and this sensitivity may affect regions of the decision boundary far beyond the local neighborhood of the point in question. This could contribute to leave-one-out unfairness, as the predictions of points in regions shifted by a training points' addition or removal will change. Adversarial training may in some cases intensify the boundaries' sensitivity to training points by penalizing inconsistent predictions in any direction within $\epsilon$ away.

Alternatively, a smoothed model returns the expected prediction over a continuous distribution centered at each point, rather than the value of the underlying model at only one point. 
While this does not remedy larger boundary changes stemming from instability, it likely does not exacerbate them, as evidenced by the effects in Figure~\ref{base_pics}.

\section{Discussion}
\label{discussion}
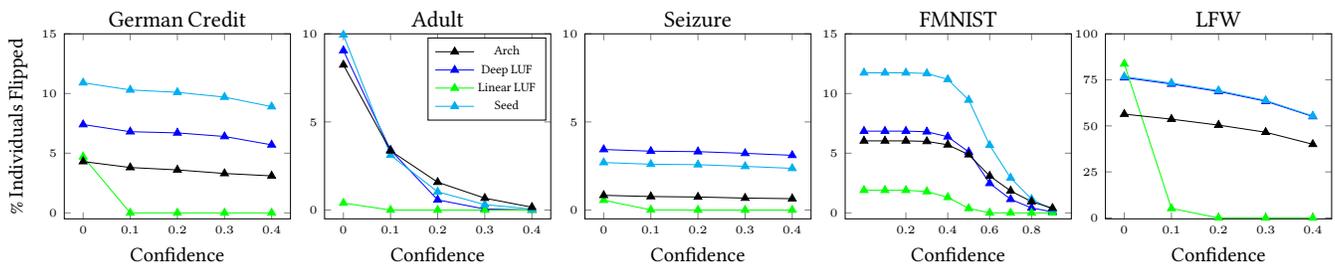
\begin{figure*}
\resizebox{\textwidth}{!}{%



\begin{tikzpicture}
\begin{groupplot}[%
	group style = {group size=5 by 1, horizontal sep=15pt, vertical sep=20pt}, 
	footnotesize,
	xtick={0, 0.1, 0.2, 0.3, 0.4},
	ymin=-0.5,
	legend style={nodes={scale=0.8}},
	height=1.75in,
	width=2.0in,
	label style={font=\normalsize},
	title style={font=\large},
	tick label style={font=\tiny}
]

\nextgroupplot[ymax=15, ytick={0, 5, 10, 15 }, ylabel={\% Individuals Flipped}
, title={German Credit}, xlabel={Confidence}]
\addplot[color=blue,mark=triangle*,blue] coordinates {
	(0.0, 7.3999999999999995)
	(0.1, 6.800000000000001)
	(0.2, 6.7)
	(0.30000000000000004, 6.4)
	(0.4, 5.7)
};

\addplot[color=green,mark=triangle*,green] coordinates {
	(0.0, 4.7)
	(0.1, 0.0)
	(0.2, 0.0)
	(0.30000000000000004, 0.0)
	(0.4, 0.0)
};

\addplot[color=cyan,mark=triangle*,cyan] coordinates {
(0.0, 10.9)
(0.1, 10.299999999999999)
(0.2, 10.100000000000001)
(0.30000000000000004, 9.700000000000001)
(0.4, 8.9)
};

\addplot[color=black,mark=triangle*,black] coordinates {
(0.0, 4.3)
(0.1, 3.8)
(0.2, 3.5999999999999996)
(0.30000000000000004, 3.3000000000000003)
(0.4, 3.1)
};


\nextgroupplot[ymax=10, ytick={0, 5, 10}, title={Adult}, xlabel={Confidence}]

\addlegendimage{black, mark=triangle*}
\addlegendentry{Arch}
\addlegendentry{Deep LUF}
\addlegendentry{Linear LUF}
\addlegendentry{Seed}

\addplot[color=blue,mark=triangle*,blue] coordinates {
	(0.0, 9.051123930773821)
	(0.1, 3.401631191565546)
	(0.2, 0.5768848219614084)
	(0.30000000000000004, 0.056362310191631855)
	(0.4, 0.0)
};

\addplot[color=green,mark=triangle*,green] coordinates {
	(0.0, 0.394536171341423)
	(0.1, 0.0)
	(0.2, 0.0)
	(0.30000000000000004, 0.0)
	(0.4, 0.0)
};

\addplot[color=cyan,mark=triangle*,cyan] coordinates {
	(0.0, 9.939659173794842)
	(0.1, 3.113188780584842)
	(0.2, 1.034414163517008)
	(0.30000000000000004, 0.30833499104833895)
	(0.4, 0.053046880180359385)
};
\addplot[color=black,mark=triangle*,black] coordinates {
	(0.0, 8.23884357801207)
	(0.1, 3.361846031430277)
	(0.2, 1.578144685365692)
	(0.30000000000000004, 0.6730322922883097)
	(0.4, 0.15914064054107818)
};


\nextgroupplot[ymax=10, ytick={0, 5, 10}, title={Seizure},
xlabel={Confidence}]

\addplot[color=blue,mark=triangle*,blue] coordinates {
	(0.0, 3.4347826086956523)
	(0.1, 3.3391304347826085)
	(0.2, 3.3130434782608695)
	(0.30000000000000004, 3.217391304347826)
	(0.4, 3.1043478260869564)
};

\addplot[color=green,mark=triangle*,green] coordinates {
	(0.0, 0.5565217391304348)
	(0.1, 0.008695652173913044)
	(0.2, 0.0)
	(0.30000000000000004, 0.0)
	(0.4, 0.0)
};

\addplot[color=cyan,mark=triangle*,cyan] coordinates {
	(0.0, 2.6956521739130435)
	(0.1, 2.6)
	(0.2, 2.5739130434782607)
	(0.30000000000000004, 2.4782608695652173)
	(0.4, 2.365217391304348)
};
\addplot[color=black,mark=triangle*,black] coordinates {
	(0.0, 0.8347826086956521)
	(0.1, 0.7652173913043478)
	(0.2, 0.7391304347826086)
	(0.30000000000000004, 0.6782608695652174)
	(0.4, 0.6434782608695652)

};


\nextgroupplot[ymax=15, ytick={0, 5, 10, 15}, xtick={0.2, 0.4, 0.6, 0.8}, title={FMNIST}, xlabel={Confidence}]

\addplot[color=blue,mark=triangle*,blue] coordinates {
	(0.0, 6.847142857142857)
(0.1, 6.847142857142857)
(0.2, 6.847142857142857)
(0.30000000000000004, 6.797142857142857)
(0.4, 6.375714285714286)
(0.5, 5.102857142857143)
(0.6000000000000001, 2.472857142857143)
(0.7000000000000001, 1.1371428571428572)
(0.8, 0.41571428571428576)
(0.9, 0.08142857142857143)
};

\addplot[color=green,mark=triangle*,green] coordinates {
	(0.0, 1.892857142857143)
(0.1, 1.892857142857143)
(0.2, 1.892857142857143)
(0.30000000000000004, 1.7871428571428574)
(0.4, 1.3014285714285714)
(0.5, 0.3757142857142857)
(0.6000000000000001, 0.0014285714285714286)
(0.7000000000000001, 0.0)
(0.8, 0.0)
(0.9, 0.0)
};

\addplot[color=cyan,mark=triangle*,cyan] coordinates {
	(0.0, 11.732857142857144)
(0.1, 11.732857142857144)
(0.2, 11.732857142857144)
(0.30000000000000004, 11.682857142857143)
(0.4, 11.18)
(0.5, 9.457142857142857)
(0.6000000000000001, 5.667142857142857)
(0.7000000000000001, 2.927142857142857)
(0.8, 1.1514285714285715)
(0.9, 0.2571428571428571)
};

\addplot[color=black,mark=triangle*,black] coordinates {
	(0.0, 6.031428571428572)
(0.1, 6.031428571428572)
(0.2, 6.031428571428572)
(0.30000000000000004, 5.988571428571428)
(0.4, 5.702857142857143)
(0.5, 4.877142857142857)
(0.6000000000000001, 3.092857142857143)
(0.7000000000000001, 1.8414285714285712)
(0.8, 0.9414285714285714)
(0.9, 0.3914285714285714)

};



\nextgroupplot[ymax=100, ytick={0, 25, 50, 75, 100 }, title={LFW},  xlabel={Confidence}]
\addplot[color=blue,mark=triangle*,blue] coordinates {
	(0.0, 76.18943692710607)
	(0.1, 72.73024879965081)
	(0.2, 68.69271060672195)
	(0.30000000000000004, 63.40026189436927)
	(0.4, 55.085115670013096)
};
\addplot[color=green,mark=triangle*,green] coordinates {
	(0.0, 83.68616324749017)
	(0.1, 5.150589262330859)
	(0.2, 0.021824530772588387)
	(0.30000000000000004, 0.0)
	(0.4, 0.0)
};

\addplot[color=cyan,mark=triangle*,cyan] coordinates {
(0.0, 76.73505019642077)
(0.1, 73.25403753819293)
(0.2, 69.1619380183326)
(0.30000000000000004, 63.814927979048456)
(0.4, 55.357922304670446)
};

\addplot[color=black,mark=triangle*,black] coordinates {
(0.0, 56.3836752509821)
(0.1, 53.64469663902226)
(0.2, 50.41466608467918)
(0.30000000000000004, 46.55172413793103)
(0.4, 39.993452640768226)
};


\end{groupplot}


\end{tikzpicture}
}
    \caption{Arbitrariness in decision outcome as a result of changes in random seed, and small changes in architecture, are presented alongside expected LUF, i.e. arbitrariness from small changes in the training set. Calculation methods are described in ~\ref{discussion}. We present these results to motivate a wider connection between learning algorithm stability and fairness, beyond LUF. Notice the difference in scale across graphs.}
    \label{gen_stab}
\end{figure*}

Our study focused on instability to changes in training data, as this type of stability is particularly well-studied due to its relevance to generalization and privacy. 
However, there are other potential sources of instability that may lead to arbitrary outcomes as well: for example, random initialization, batching order, and model architecture.
If a difference in any of these choices results in a difference in outcome for an individual---e.g., if a change in random initialization frequently leads to a change in predicted credit risk for someone---then this too could be seen as unfair, as it would call into question the robustness of any supposed justification.

To establish a preliminary understanding of the degree to which these sources introduce changes in outcome similar to LUF,
we experimentally investigate the percentage of changed outcomes resulting from varying the random seed prior to initializing and training models, as well as from the choice of model architecture. 
Figure~\ref{gen_stab} shows these results for all of the datasets studied in Section~\ref{loo_exp_base}, alongside the corresponding measurements for LUF.
The experimental setup largely follows that described in Section ~\ref{loo_exp_base}. We isolate the effect of each potential variable causing instability unfairness (architecture, random seed, and leave-one-out unfairness) in its own experiment; keeping other sources of instability controlled. For the random seed experiments, we train the same model with 100 different random seeds and calculated the effects of instability in the same manner as calculating LUF described in Section ~\ref{loo_exp_base}; for the experiments calculating the fairness effects of changes in architecture, we train the model on three different architectures, as described for the experiments verifying consistency in LUF in Section~\ref{loo_exp_base}. Further information on the architectures considered can be found in the supplementary material.

As Figure~\ref{gen_stab} shows, any of these aspects in a model can affect model behavior over a substantial percentage of the overall dataset. Interestingly, LUF seems to have a more consistent effect across points with high prediction confidence than arbitrariness resulting from a change in architecture. LUF seems to have a similar effect to changing random seed and initialization, as changing seed produces a larger effect in FMNIST and German Credit, but a smaller effect in Adult and Seizure models.
While these other sources of instability unfairness are interesting avenues for future work, we focus on leave-one-out unfairness in this paper due to its useful connections to other areas of the machine learning literature, bridging the fields of fairness to those of stability and privacy as discussed in Section ~\ref{loo_unfairness_def}, and also to the field of robustness, as explored in Section ~\ref{adversarial_exp}.



\section{Related Work}
\label{related}


Leave-one-out unfairness views the problem of learning instability~\citep{bonnans2013perturbation,bousquet2002stability} from a fairness perspective. While deep learning is generally understood not to enjoy strong stability properties, our results are among the few systematic studies of the extent, and potential ramifications, of their instability. 
Hardt et al. show that even nonconvex models trained using Stochastic Gradient Descent remain stable over a small number of iterations, and that popular heuristics like dropout and $\ell_2$ regularization help~\cite{hardt2016trainfaster}, and provide some experimental demonstrations. Towards achieving stability in deep learning, Kuzborskij et al.~\citep{Kuzborskij2018DataDependentSO}, develop a screening protocol for choosing random initalizations that improve stability.


\textit{Memorization}, as defined by Feldman~\cite{feldman2019shorttale}, is a symptom of model instability where a model predicts the correct output on a given point if it is in the training set, and incorrectly otherwise. There has been much recent work unearthing the potential for
memorization in deep neural networks~\citep{zhang2016understanding}, discussion about the extent of the phenomenon in practice~\citep{arpit2017closer} as well as arguments for its usefulness~\cite{feldman2019shorttale}. Memorization is closely related to leave-one-out unfairness in it is a measure of stability, and crucially, focuses on how instability affects a given point, rather than an average. However, leave-one-out fairness is much broader than memorization. Memorization quantifies how much removing a given point from the training set affects that whether that particular point is predicted correctly. Leave-one-out fairness quantifies how the consistency, not the error, of a given point's prediction is affected by \textit{any other point}.

A well-known meeting point of stability and privacy is differential privacy~\citep{dwork2006differential}, which quantifies privacy risk in terms of a uniform, information-theoretic notion of stability. Leave-one-out fairness is related to, but weaker than, differential privacy, as shown in Section~\ref{loo_unfairness_def}. 
Instability also worsens concrete privacy attacks: oversensitivity to the training set can affect a model's parameters, which can be leveraged to perform membership inference~\citep{yeom2018privacy, leino2019memorizationprivacy, song2017remembertoomuch}. 
Our experiments in Section ~\ref{adversarial_exp} may suggest that this phenomenon has a connection to leave-one-out unfairness, in that adversarial training increases both LUF and the potential for membership inference attacks~\citep{yeom2020overfitting,song2019membership}.

There is little work that connects \textit{fairness} and  stability. Leave-one-out fairness is an individual-based fairness notion. While there are several definitions of ``individualized'' fairness~\cite{dwork2012fairness, dwork2018fairness, joseph2016fairness,kusner2017counterfactual}, they are rarely operationalized in common fairness testing platforms, as they can be difficult to calculate. 
In addition to already-noted differences from prior notions of fairness, expected LUF can be effectively measured on real datasets to give insight into whether an individual may be subject to unfair treatment at inference time.

\section{Conclusion}
We present \textit{leave-one-out} fairness, a connection between algorithmic stability and fairness. We demonstrate the extent to which deep models are leave-one-out unfair, and experimentally showed that this behavior does not depend on generalization error. Interestingly, adversarial training worsens leave-one-out unfairness in deep models, while random smoothing often mildly mitigates it, showing that leave-one-out fairness is not dependent on robustness or individual fairness. These results may suggest an interesting geometric intuition of deep networks' sensitivity to their training points. Finally, we note that LUF may be undesirable in sensitive applications, as it casts doubt on the justifiability of a model's decision. 

\subsection*{Acknowledgments} This paper is based on work supported by the National Science Foundation under Grants No. CNS-1943016 and CNS-1704845.

\newpage

\bibliography{bib}
\bibliographystyle{plainnat}

\clearpage
\newpage

\appendix

\section{Proofs}

We present the full proofs from Section 2. 

\begin{proposition}
Let $h$ be a learning rule optimizing 0-1 loss and $\epsilon(m)$ be a montonically-decreasing function such that $\mathrm{LUF}(h,S,x) \le \epsilon(n)$ for all $S\sim\mathcal{D}^m$ and $x$. Then $h$ is on-average leave-one-out stable with rate at most $\epsilon(m)$.
\begin{proof}

We prove the case for binary classification. The result generalizes to multiclass problems in a straighforward fashion.
Note that because LUF is bounded for all $S$, we can disregard the expectation over $S$ in the definition of LOO-stability, and assume that the randomness in the expectations comes from the learning rule $h$ exclusively. By linearity of expectation, we have that, 
\begin{align*}
\E[|\ell(h_{S}, z_i)-\ell(h_{S^{(\setminus i)}}, z_i)|] 
& = \Pr[h_{S}(z_i) \ne h_{S^{(\setminus i)}}(z_i)] \\
& = |\Pr[h_{S}(z_i)=1] - \Pr[h_{S^{(\setminus i)}}(z_i) = 1]|
\end{align*}

Now, assuming, 
$$\forall x, S, \max_{i}|\Pr[h_S(x)=1] - \Pr[h_{S^{(\setminus i)}}(x)=1]| \le \epsilon(m)$$ we have: 
\begin{align*}
\frac{1}{m}\sum_{i=1}^m\E_{\substack{S \sim \D^n}}[|\ell(h_{S}, z_i)-\ell(h_{S^{(\setminus i)}}, z_i)|] \\
\leq \max_S\frac{1}{m}\sum_{i=1}^m \E[|\ell(h_{S}, z_i)-\ell(h_{S^{(\setminus i)}}, z_i)|]
\end{align*}
The result follows by noting that each term in the above sum is bounded by $\epsilon(m)$.

\end{proof}
\end{proposition}

\begin{proposition}
Let $h$ be an $(\epsilon, \delta)$-differentially private learning rule, and $x\sim\mathcal{D}$ be a point. Then
$
\mathrm{LUF}(h, x) \le e^\epsilon - 1 + \delta
$.
\begin{proof}
We prove the case where $h$ produces binary classifiers. The extension to multi-class learning is straightforward. The result follows from a general property of differentially-private algorithms~\citep[Lemma~6]{Dwork15Adaptive} which is that when $h_S$ ranges in $[0,1]$,
\[
|\mathbb{E}[h_S(x)]-\mathbb{E}[h_{S^{(\setminus i)}}(x)]|  \le e^\epsilon-1+\delta
\]
Noting that $\mathbb{E}[h_S(x)] = \Pr[h_S(x) = 1]$, the result follows.






 
\end{proof}
\end{proposition}

\section{Additional LFW Results}
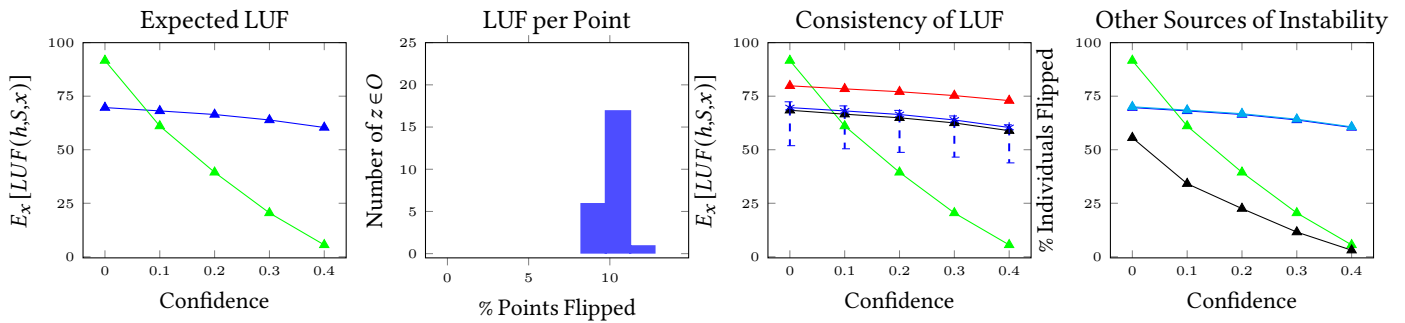
\begin{figure*}
	\centering



\begin{tikzpicture}
\begin{groupplot}[%
	group style = {group size=4 by 1, horizontal sep=30pt},
	footnotesize,
	yticklabel style={
		/pgf/number format/precision=3,
		/pgf/number format/fixed
	},
	ymin=-0.5,
	legend style={nodes={scale=0.8}},
	height=1.75in,
	width=2.0 in,
	label style={font=\normalsize},
	title style={font=\large},
	tick label style={font=\tiny}
]

\nextgroupplot[ymax=100, ytick={0, 25, 50, 75, 100}, title={Expected LUF}, xlabel={Confidence}, ylabel={$E_x[LUF(h,S,x)]$}]

\addplot[color=blue,mark=triangle*,blue] coordinates {
	 (0.0, 69.67481449148843)
	(0.1, 68.12527280663467)
	(0.2, 66.44478393714535)
	(0.30000000000000004, 63.8913138367525)
	(0.4, 60.38847664775208)
};
\addplot[color=green,mark=triangle*,green] coordinates {
	(0.0, 91.64120471409865)
(0.1, 61.065037101702316)
(0.2, 39.43692710606722)
(0.30000000000000004, 20.47140986468791)
(0.4, 5.554343081623745)
};

\nextgroupplot[ymax=25, ylabel={Number of $z\in O$}, title={LUF per Point}, xtick={0, 5, 10, 15}, ybar=-6pt, xlabel={\% Points Flipped}]


\addplot[
    draw=none, 
    fill=blue,
    fill opacity=0.7, xtick={0, 5, 10, 15},
  ] coordinates {

   (0.0, 0.0)
(1.5, 0.0)
(3.0, 0.0)
(4.5, 0.0)
(6.0, 0.0)
(7.5, 0.0)
(9.0, 6.0)
(10.5, 17.0)
(12.0, 1.0)
(13.5, 0.0)};

\nextgroupplot[ymax=100, ytick={0, 25, 50, 75, 100}, title={Consistency of LUF}, xlabel={Confidence}, ylabel={$E_x[LUF(h,S,x)]$}]

\addplot[color=blue,mark=triangle*,blue] coordinates {
};
\addplot[color=green,mark=triangle*,green] coordinates {
	(0.0, 91.64120471409865)
(0.1, 61.065037101702316)
(0.2, 39.43692710606722)
(0.30000000000000004, 20.47140986468791)
(0.4, 5.554343081623745)
};
\addplot[color=black,mark=triangle*,black] coordinates {
	(0.0, 68.4308162374509)
(0.1, 66.59755565255348)
(0.2, 64.91706678306416)
(0.30000000000000004, 62.516368398079436)
(0.4, 58.89349628982977)


};

\addplot[color=red,mark=triangle*,red] coordinates {
	(0.0, 79.84504583151462)
(0.1, 78.41553906591008)
(0.2, 77.08424268878218)
(0.30000000000000004, 75.28371890004365)
(0.4, 72.9375818419904)

};

\addplot [blue, mark=asterisk, error bars/.cd, y dir=both, y explicit,
      error bar style={line width=0.75pt,dashed},
      error mark options={line width=0.25pt, mark size=2pt,rotate=90}]
    table [x=x, y=y, y error plus index=2, y error minus index=3]{%
      x   y                  y-err y-err2
     0.0 69.67481449148843 2.706241815800965 17.78699257965954
0.1 68.12527280663467 2.3461370580532304 17.688782191182902
0.2 66.44478393714535 1.9205587079877802 17.677869925796593
0.30000000000000004 63.8913138367525 1.8878219118289081 17.33958969882147
0.4 60.38847664775208 1.2549105194238237 16.53208206023571
      
    };

\nextgroupplot[ymax=100, ytick={0, 25, 50, 75, 100 }, title={Other Sources of Instability},  xlabel={Confidence}, ylabel={\% Individuals Flipped}]
\addplot[color=blue,mark=triangle*,blue] coordinates {
(0.0, 69.67481449148843)
	(0.1, 68.12527280663467)
	(0.2, 66.44478393714535)
	(0.30000000000000004, 63.8913138367525)
	(0.4, 60.38847664775208)
};
\addplot[color=green,mark=triangle*,green] coordinates {
(0.0, 91.64120471409865)
(0.1, 61.065037101702316)
(0.2, 39.43692710606722)
(0.30000000000000004, 20.47140986468791)
(0.4, 5.554343081623745)
};

\addplot[color=cyan,mark=triangle*,cyan] coordinates {
(0.0, 70.05674378000873)
(0.1, 68.49628982976867)
(0.2, 66.81580096027936)
(0.30000000000000004, 64.25141859450022)
(0.4, 60.672195547795724)
};

\addplot[color=black,mark=triangle*,black] coordinates {
(0.0, 55.58707987778263)
(0.1, 34.144478393714536)
(0.2, 22.512003491924922)
(0.30000000000000004, 11.545176778699258)
(0.4, 3.077258838934963)
};

\end{groupplot}
\end{tikzpicture}
    \caption{Additional experiments on the LFW dataset, with identical setup to the results presented in the paper, but with a static group of face pairs encountered during training. From left to right, we have: the experiments presented in Section 5 concerning the extent of LUF in deep models (first two graphs), experiments from Section 5 showing the consistency of leave-one-out unfair behavior across different model architectures and seeds, and experiments showing the effect of instability from other sources discussion in Section 6.}
    \label{decision_bound}
\end{figure*}
We present additional experiments on the LFW dataset, with identical setup to the results presented in the paper, with the exception of the training set face pair generation process.  In this setup, the model is trained on a static set of face pairs common across all models, as opposed to a being trained with a generator creating random face pairs that may differ on each training run. This static setup along with the rest of the precautions taken in all our experiments ensures that all possible sources of instability are controlled, aside from leave-one-out unfairness. One other  difference in this set of experiments is that we sample 25 points to remove from the dataset, as opposed to 50 as in the results presented in the main paper. The results are qualitatively similar to the results presented in the paper, and still show far greater expected LUF than any other models presented, with LUF of approximately 69\%. However, due to memory constraints, a comparatively small set of pairs of faces from LFW can be contained statically in memory, and the accuracy of the model suffers: the accuracy of $h_S$ in this setup is 76\%, with a generalization error of 22\%. 

\section{Datasets}
\label{app_data}
The German Credit data set consists of individuals financial data, with a binary response indicating their creditworthiness. There are 1000 points, and 20 attributes. We one-hot encode the data to get 61 features, and standardize the data to zero mean and unit variance using SKLearn Standard scaler. We partitioned the data intro a training set of 700, a leave-one-out-set of 100, and a test set of 200.  

The Adult dataset consists of a subset of publicly-available US Census data, binary response indicating annual income of $>50$k. There are 14 attributes, which we one-hot encode to get 96 features. We normalize the numerical features to have values between $0$ and $1$. After removing instances with missing values, there are $30,162$ examples which we split into a training set of 14891, a leave one out set of 100, and a test set of 1501 examples. 

The Seizure dataset comprises time-series EEG recordings for 500 individuals, with a binary response indicating the occurrence of a seizure. This is represented as 11500 rows with 178 features each. We split this into 7,950 train points and 3,550 test points. We standardize the numeric features to zero mean and unit variance. 

Fashion MNIST contains images of clothing items, with a multilabel response of 10 classes. 
There are 60000 training examples and 10000 test examples. We pre-process the data by normalizing the numerical values in the image array to be between $0$ and $1$.

The Labeled Faces in the Wild dataset (LFW) consists of 13,000 cropped images of 1,680 individuals' faces, with a multiclass label of 1,680 classes, corresponding to which individual is in what image. We pre-process the images by normalizing the numerical values in the image array to be between $0$ and $1$. 
Since the model that we use on the data is a face-matching model, we create a training set of pairs of images from the processed LFW. First, we split the original LFW dataset in a training set and test set, of sizes 6,873 and 2,291. 
For the results presented in the main paper, we use a data generator to create 6,873 pairs of images from the training set on each epoch. These pairs of images have a 50\% match rate (that is, 50\% of the pairs are of the same individual, and 50\% are not).
For the results for LFW presented in the supplementary material, we generate a static training set of 6,873 face pairs that stay consistent epoch to epoch. Note that this results in many fewer unique face pairs seen by the face-matching algorithm. 
For the test set in both the main paper and the supplementary material, we generate a static 2,291 pairs of images from the test set, again with 50\% match rate.

\section{Calculating LUF in All Experiments}
\label{how_to_calc}
We provide a description of how we calculated (an approximation of) LUF in our experiments.
Given predictions of the entire dataset for both $h_{S}$ and $h_{S^{(\setminus i)}}$ models: For binary classification models, for each $h_{S^{(\setminus i)}}$, and for $h_{S}$, we calculate whether the output is class 1 or 0. We then take the difference in binary predictions from the baseline model and $h_{S^{(\setminus i)}}$, for each of the 100 $h_{S^{(\setminus i)}}$. We choose the maximum difference over all $h_{S^{(\setminus i)}}$ for each point (i.e., searching to see if the removal of \emph{any} point removed results in a change in prediction for an individual in the distribution.) This approximates the leave-one-out unfairness for each $x$ in the dataset, in the setting of a deterministic learning rule, as described in the main paper. Note that the approximation arises from the fact that we sample 100 points at random with which to create $h_{S^{\setminus i}}$, rather than creating a different model for each point in the dataset. We then divide the number of individuals experiencing LUF by the size of the dataset to calculate the expected LUF over the dataset.

For multiclass problems, we follow a similar procedure, except that we calculate the probabilities that $h_{S^{(\setminus i)}}$ and $h_{S}$ output a given class for each class, compute the differences between these probabilities, take the maximum over $k$ classes, and then proceed as in the binary case.

Finally, for calculating the effects of random seed and architecture on unfair arbitrariness as displayed in the discussion, we follow the exact same procedure as above, but where $h_{S^{(\setminus i)}}$ is a model trained on a different seed, or in the architecture results, on a different architecture.

\section{Experimental Setup Further Details}
\label{exp_setup_appendix}

For German Credit and Seizure datasets, we trained all models in the paper with three hidden layers, of size 100, 32, and 16, over 100 epochs. The inner activations are ReLu, and the final activation is Sigmoid. The model is trained with binary crossentropy. We used the Adam optimizer with the default parameters used by Keras. The linear models for both datasets are trained over 100 epochs with a batch size of 32. For the German Credit models in Section 3, and the random smoothing experiments in Section 4, we use a batch size of 32. For the adversarially trained models, we use a batch size of 4. For the models used to compare the variance of LUF over different architectures, we train a one hidden layer of size 100 with the same hyperparameters as described for the main model, and a 3-hidden layer model with layer sizes 64, 16, and 8, again with the other hyperparamters kept constant.

For the Adult dataset, our main model was one hidden layer of size 200, over 50 epochs with a batch size of 128. The activations, loss, and optimizer were the same as those for German Credit. The linear models for Adult were also trained with a batch size of 128 over 50 epochs. For the adversarial experiments, we use a batch size of 32.  For the models used to compare the variance of LUF over different architectures, we train a one hidden layer of size 100 with the same hyperparameters as described for the main model, and a 3-hidden layer model with layer sizes 128, 32, and 16, again with the other hyperparamters kept constant.

For the Seizure dataset, we trained the main models in the paper with three hidden layers, of size 128, 32, and 16, over 100 epochs. All models, including the linear models, were trained with a batch size of 128. The activations, loss, and optimizer were the same as those for German Credit. For the models used to compare the variance of LUF over different architectures, we train models with the same architecture as the German Credit datasets, with the rest of the parameters kept the same as in the main experiments.

For the FMNIST data set, for all models in the paper, we used a LeNet\citep{lenet} architecture modified for the size of the data, trained with dropout: this consists of 2 convolutional layers with 20 and 50 channels respectively, each followed by a max pooling layer, and finally a dense layers with 200 neurons. We train with SGD, batch size 128, and 50 epochs. For the linear data, we used a batch size of 128 and 50 epochs as well. For the adversarially trained models, we use a batch size of 32.
The models over FMNIST with varying architecture included a the same model described above, but trained without dropout, and a shallower model with the middle layer (along with the corresponding pooling and convolution layers) removed.

For LFW, we train a face-matching model, that takes a pair of images and outputs a binary label whether or not the pair of images are of the same individual. The LFW face-matching model consists of a concatenation layer composing the two input images, a 4-layer convolutional stack, followed by a dense layer, and a sigmoid output. It is trained with the Adam optimizer with the default learning rate, batch size of 128, over 50 epochs. To compare the effect of LUF across architectures, we train use a ResNet50 model, pre-trained on ImageNet weights from Keras, modified to take two images as input and have a Sigmoid output. We also compare the effects of LUF on a model with the same architecture as the original one described, but with doubled filter sizes for the convolutions. All other models are trained with the same hyperparameters as the original model.

\section{Experimental Setup For Decision Boundary Images}
\label{imgs_appendix}
To generate the pictures in Figure 1, we train a model 3 Relu layers, each with 1000 neurons, trained on 100 uniform-random points with Bernoulli labels.

\label{appendix}

\end{document}